\documentclass[10pt,journal,compsoc]{IEEEtran}
\usepackage{lineno,hyperref}
\usepackage{amsmath,amssymb,amsfonts}
\usepackage{algorithmic}
\usepackage{graphicx}
\usepackage{textcomp}
\usepackage{xcolor}
\usepackage{color,colortbl}
\usepackage{algorithm}
\usepackage[vlined, ruled, algo2e]{algorithm2e}
\usepackage{subcaption}
\usepackage{multirow}
\usepackage{caption}
\usepackage{float}
\usepackage{stfloats}
\usepackage{comment}
\modulolinenumbers[5]
\ifCLASSOPTIONcompsoc
  \usepackage[nocompress]{cite}
\else
  \usepackage{cite}
\fi
\ifCLASSINFOpdf
\else
\fi
\begin{document}
%
\title{WiFi Fingerprint Clustering for Urban Mobility Analysis}
%
%
%
%

\author{Sumudu Hasala~Marakkalage,
        Billy Pik Lik~Lau,
        Yuren~Zhou, Ran~Liu,
        Chau~Yuen, Wei Quin~Yow,
        and~Keng Hua~Chong
\IEEEcompsocitemizethanks{\IEEEcompsocthanksitem S.H. Marakkalage, B.P.L. Lau, Y. Zhou, R.Liu, and C. Yuen are with the  Engineering Product Development Pillar, Singapore University of Technology and Design (SUTD), Singapore.\protect\\
Corresponding author email: marakkalage@alumni.sutd.edu.sg
\IEEEcompsocthanksitem W.Q. Yow is with Humanities, Arts and Social Sciences Pillar, SUTD
\IEEEcompsocthanksitem K.H. Chong is with Architecture and Sustainable Design Pillar, SUTD.}
}

%
%


\IEEEtitleabstractindextext{%
\begin{abstract}
In this paper, we present an unsupervised learning approach to identify the user points of interest (POI) by exploiting WiFi measurements from smartphone application data. Due to the lack of GPS positioning accuracy in indoor, sheltered, and high rise building environments, we rely on widely available WiFi access points (AP) in contemporary urban areas to accurately identify POI and mobility patterns, by comparing the similarity in the WiFi measurements. We propose a system architecture to scan the surrounding WiFi AP, and perform unsupervised learning to demonstrate that it is possible to identify three major insights, namely the indoor POI within a building, neighborhood activity, and micro mobility of the users. Our results show that it is possible to identify the aforementioned insights, with the fusion of WiFi and GPS, which are not possible to identify by only using GPS.
\end{abstract}

}

\maketitle

\IEEEdisplaynontitleabstractindextext

\IEEEpeerreviewmaketitle

\IEEEraisesectionheading{\section{Introduction}\label{sec:introduction}}

In recent times, mobile crowdsensing (MCS) has obtained a huge attention due to the pervasiveness of smart mobile devices, their in-built sensing abilities, and the fact that they have become an everyday carry item by humans. Therefore, plethora of MCS applications have become prominent in various sectors, namely transportation ~\cite{farkas2015crowdsending}, healthcare~\cite{leonardi2014secondnose}, and social networking platforms~\cite{hu2014multidimensional}. A particular phenomenon can be monitored by diverse information harnessed through smartphone applications with proper crowd participation \cite{hoteit2014estimating, kang2013exploring}. In mobility analysis applications, identifying detailed motion pattern information (outdoor and indoor) provides comprehensive insights on user mobility \cite{li2008mining,lou2009map,gamanayake2020cluster,helgason2013opportunistic}. Knowing the user points of interest (POI) is paramount in mobility tracking applications to provide context-aware services. Motion pattern learning and anomaly detection of human trajectories is done in \cite{suzuki2007learning} using Hidden Markov Models. Past research has conducted to detect the type of environment (i.e. indoor and outdoor) with the fusion of smartphone based sensor data \cite{zhou2012iodetector, 7917558, Shin2012Unsupervised}. Understanding the elderly lifestyle is studied in \cite{marakkalage2018understanding}, using smartphone application data. Its main focus is to extract regions of interest (ROI) and POI with sensor fusion. Nonetheless, in contemporary urban indoor places (e.g. shopping malls, apartment complexes etc.), where massive crowd movements happen, aforementioned work are not sufficient to identify the indoor POI granularity. 

Mobility tracking in indoor environments is a challenge because of the constraints to acquire fine-grained location based information in such places. Especially in high rise urban buildings/apartments, it is difficult to identify when people leave their home/office, and visit common areas within the same building or neighborhood POI, by only using GPS data. Even the GPS accuracy is low in those scenarios, we can distinguish such different places by incorporating WiFi data. Urban environments nowadays are equipped with plentiful of WiFi access points (AP). Hence, by combining or fusing GPS and WiFi information we intend to identify indoor POI (as first introduced in our previous work~\cite{marakkalage2019identifying}, and improved POI extraction technique in this paper), and introducing neighborhood activities, and micro mobility analysis information in this paper, by utilizing crowdsensing smartphone data. 
Prior research has utilized WiFi AP information to generate indoor floorplans \cite{Shin2012Unsupervised, alzantot2012crowdinside} and to identify indoor indoor locations through localization~\cite{zhu2014spatio, liu_crowdsensing_2019,liu_ieee_sensors2017,liu2019collaborative,tian2019rf}.
Major drawback of those work is, they require data collection in high sampling rates, which incurs high power consumption (a prime challenge in MCS \cite{ganti2011mobile, lau2019survey, wu2014smartphones, marakkalage2019real}).
Furthermore, an extensive labor cost is required when creating indoor fingerprint maps, which is another drawback.

To identify the indoor POI, we focus on the mobility pattern of a typical user in indoor environments like shopping mall or apartment complex with POI, where users frequently visit, yet it is challenging to identify such POI by only using GPS location data, due to the lack of accuracy in indoor environments. For a particular user, after processing the raw GPS data, they may get clustered into one POI, when he visits a particular shopping mall, but in reality the user may have visited multiple POIs (e.g. visit different shops) within the same mall. This is due to the two dimensional nature of GPS data, which limits differentiation between multiple indoor POIs.
Therefore, fusing GPS with WiFi data helps to identify such indoor POI. 

Neighborhood activity analysis is conducted to understand the POI, where users visit in their residence neighborhood (e.g. common areas in an apartment complex). A user may visit a convenient store at downstairs for grocery, or may visit a common area in the same building to mingle with friends, as getting a short break while staying at home. Since those apartment buildings are high rise buildings (e.g. in Singapore, most of the apartments are high rise multi-storey buildings), such vertical mobility may not reflect in GPS location tracking. Hence, exploiting surrounding WiFi AP information is useful when identifying such neighborhood activities.

Micro mobility analysis is conducted to understand the mobility patterns of the users, due to blockage of GPS signal in sheltered walkway or void deck under high rise building. GPS alone may not give accurate information on such scenarios. Therefore, it would misinterpret same physical location with fluctuated GPS locations due to lack of accuracy. Investigating the surrounding WiFi AP information would indicate those fluctuated locations as one location, due to the similarity of WiFi measurement.

In a nutshell, the three main objectives of this article are to understand the distinct POI in indoor environments visited by users, neighborhood activity analysis, and micro mobility analysis. We verify the effectiveness of the proposed method, based on crowdsensing data collected from volunteers along with the visited POI ground truth. The contributions in this paper are listed below.

\begin{itemize}
	\item Introducing an unsupervised method to make use of the similarity of surrounding WiFi AP information of users to understand their mobility, and verify with real-world collected data. 
	\item Clustering of WiFi fingerprint in a given GPS POI to identify the distinct WiFi based POI of users in an indoor environment, the revisited POI by the same set of users, and the common POI among users.
	\item GPS and WiFi data fusion to identify the neighborhood activity and heat map by excluding stay home duration.
	\item Clustering of travel path WiFi fingerprints to identify the neighborhood micro mobility patterns that move under covered walkway or cutting across buildings.
\end{itemize}

The rest of this paper is organized as follows. In Section \ref{sec_2}, the proposed system and its overview is presented. In Section \ref{sec_3}, the unsupervised POI extraction technique and technical evaluations are presented. In Section \ref{sec_4}, the neighborhood activity analysis process is presented along with the results. In Section \ref{sec_5}, the micro mobility analysis technique and the results are presented. Section \ref{sec_6} presents the discussion and future work to conclude the paper.  

\section{System Overview}
\label{sec_2}
\begin{figure}[!htb] 
	\centering
	\includegraphics[width=0.49\textwidth]{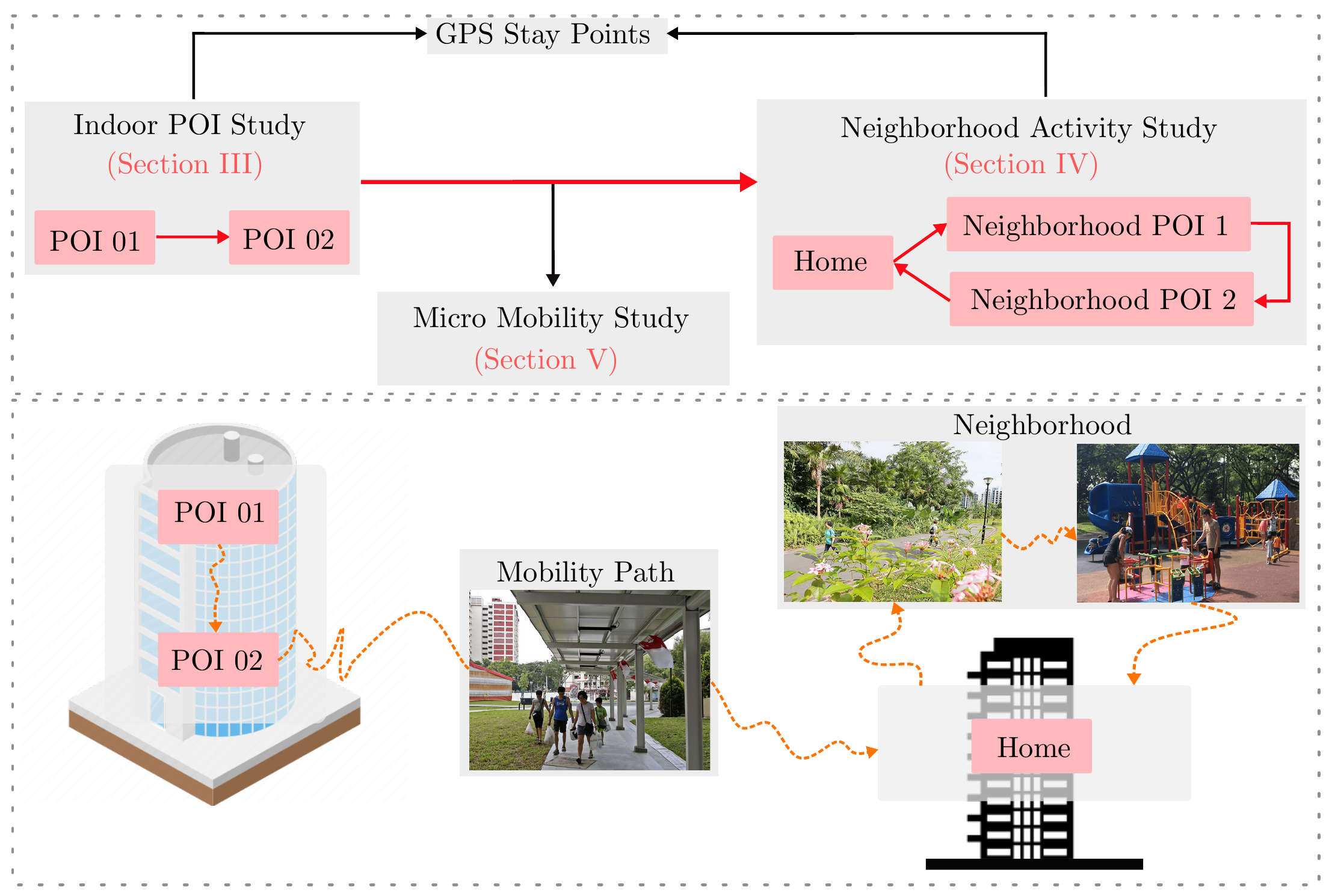} 
	\caption{Example of a user's trajectory}
	\label{fig:trajectory} 
\end{figure}

Identifying the trajectory of a user is essential in mobility analysis. Figure \ref{fig:trajectory} shows a sample trajectory of a user. It consists of GPS stay points, indoor POI within a GPS stay point, neighborhood activity happen during a GPS stay point time duration, but doesn’t capture due to low GPS accuracy
in indoor/high rise urban environments, and micro mobility (link) between two GPS stay points. In this paper, we identify those three insights on such a user trajectory.

\begin{figure}[!htb] 
	\centering
	\includegraphics[width=0.49\textwidth]{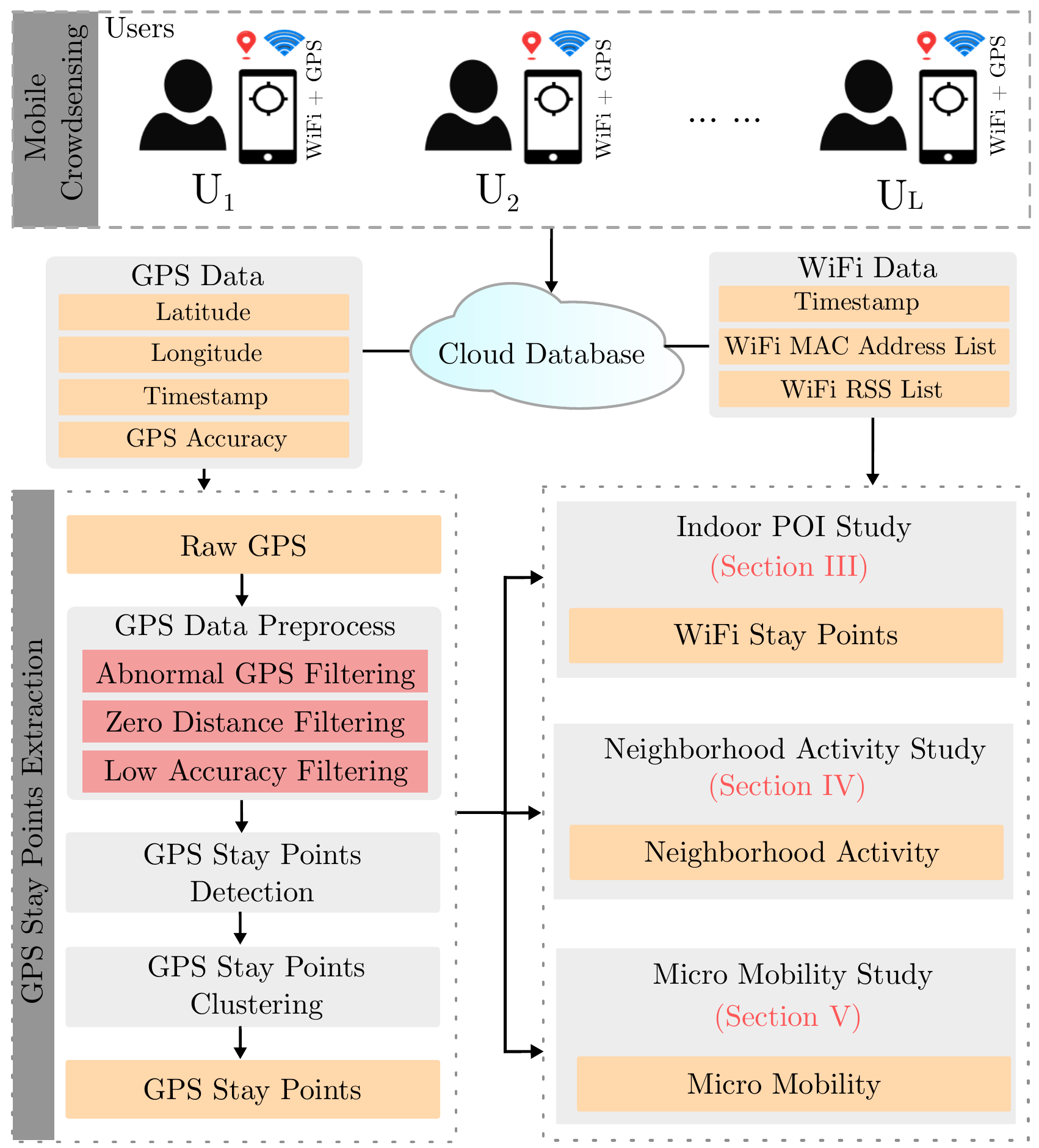} 
	\caption{Overview of the proposed system} 
	\label{fig:overview} 
\end{figure}

The proposed system comprises a smartphone application (front-end) to GPS location data, and surrounding WiFi AP information, which are transferred to a cloud-based server application (back-end). The collected raw GPS and WiFi data are further processed to identify the indoor POI, neighborhood activities, and neighbourhood micro mobility patterns of the users. Figure \ref{fig:overview} shows an overview of the proposed system.

\subsection{Data Collection Mobile Apps}
The surrounding WiFi AP information, namely MAC address and corresponding received signal strength (RSS) are scanned by the smartphone application, which acquires data at a sampling rate of $5$ minutes to conserve the power, since excessive scanning of WiFi and GPS heavily impacts on mobile phone battery consumption, according to the Android API \cite{wifiscan}. 

\subsubsection{WiFi Scanning}
\label{lab:wifi}
Let MAC address of the WiFi AP be $m$, and RSS of the AP be $r$ in $dBm$. The list of surrounding AP MAC addresses and their corresponding RSS, which is also called scan result ($s$) is shown in the Equation \ref{eq_scanResult}, where $n$ is the number of AP observed in a given scan result.

\begin{equation}
\label{eq_scanResult}
s = {\{m_1,r_1\},\{m_2,r_2\},...,\{m_n,r_n\}}
\end{equation}

Each scan result and the corresponding timestamp ($t$) of the WiFi scan is stored in a list of scan results ($S$), denoted as shown in Equation \ref{eq_listOfScanResults}, where $m$ is the number of scan results in $S$. 

\begin{equation}
\label{eq_listOfScanResults}
S = {\{s_1,t_1\},\{s_2,t_2\},...,\{s_m,t_m\}}
\end{equation}

The scanned list of scan results is stored locally in the device until it is uploaded to back-end for further analysis.

\subsubsection{Data Compression}
To avoid the extensive cost in transmitting the raw data into the back-end, we compress the raw data. As shown in Table \ref{tab_compression}, we select $6$ hour duration as the upload interval as it has a significant reduce in size when compressed. Data upload happens only when the device is connected to a WiFi network. Otherwise, the smartphone application keeps the data until it connects to a network.

\begin{table}[!htb]
	\centering
	\caption{Comparison of data size before and after compression}
	\label{tab_compression}	
	\begin{tabular}{|c|c|c|}
		\hline
		\textbf{\begin{tabular}[c]{@{}c@{}}Duration of data \\ (Hours)\end{tabular}} & \textbf{\begin{tabular}[c]{@{}c@{}}Size (uncompressed)\\ (Bytes)\end{tabular}} & \textbf{\begin{tabular}[c]{@{}c@{}}Size (compressed)\\ (Bytes)\end{tabular}} \\ \hline
		0.5                                                                  & 20,701                                                                   & 656                                                                   \\ \hline
		1                                                                    & 41,401                                                                   & 791                                                                   \\ \hline
		3                                                                    & 172,501                                                                  & 1,562                                                                  \\ \hline
		6                                                                    & 345,001                                                                  & 2,565                                                                  \\ \hline
	\end{tabular}
\end{table}
\subsubsection{User Information}
Users and their smartphone details (which are used for the later part of experimental study) are shown in the Table \ref{tab_smartphones}. 

\begin{table}[!htb]
	\centering
	\caption{Users and smartphone models}
	\label{tab_smartphones}
	\begin{tabular}{|c|l|}
		\hline
		\textbf{User} & \multicolumn{1}{c|}{\textbf{Model}} \\ \hline
		A             & OnePlus 3                                  \\ \hline
		B,F             & Samsung S8                                \\ \hline
		C             & Sony Z3                                \\ \hline
		D             & Google Pixel 2                                \\ \hline
		E             & Huawei Nova 2i                             \\ \hline
		G             & Xiaomi Max 2                             \\ \hline		
		H             & Oppo F5                          \\ \hline		
		I             & Oppo R11                          \\ \hline		
		J             & Xiaomi Mix 2                         \\ \hline		
		K             & Samsung A8                         \\ \hline
		L			  & LG V30 \\ \hline		
	\end{tabular}
\end{table}
\subsection{GPS Stay Points Extraction}
\label{sec:cloud}
The received data in back-end, are processed to understand the indoor POI, neighborhood activity, and micro mobility patterns of the users.

We obtain the GPS based stay points of users, using the data processing pipeline as shown in Figure~\ref{fig:overview}'s GPS stay points extraction module. 
First, we conduct the data processing of the raw GPS data, which includes components such as removing abnormal, zero distance sequence, and low accuracy GPS location \cite{marakkalage2019real}.
The abnormal data here includes GPS data with sudden location shift within a short period of time, which can distort the actual path traveled by the users.
Zero distance often occurs when the GPS does not receive any signal from the satellite, which causes exact same location for subsequent data. 
This does not provide any meaningful data for us, and hence we filter it out.
The next technique is accuracy filtering, where low accuracy GPS data are removed, that causes high uncertainty in determining the actual location of users.
Next, we perform GPS stay point extraction~\cite{Zheng2009Mining, lau2017extracting} to obtain the list of POIs from a particular user with the timestamp for each visit.
Afterwards, we cluster POIs based on their geographical location using DBSCAN to group POI for similar places.
This briefly explains the GPS stay points extraction method. 
After obtaining the raw GPS stay points, we use the duration of the GPS stay points to further detect indoor POI for the users.

\section{Indoor POI Study}
\label{sec_3}
In this section, we present the techniques that used to extract indoor POI of the users, by processing the GPS and WiFi data collected through the smartphone based mobile application. Indoor POI extraction is performed by clustering the WiFi fingerprints and matching the corresponding cosine similarity scores. Table \ref{tab_wificluster_parameters} shows the symbols used in this section, and their description for the convenience of the reader.
\begin{table}[!htb]
	\centering
	\caption{Symbols and their description for clustering algorithm}
	\label{tab_wificluster_parameters}
	\fontsize{9pt}{9pt}\selectfont	
	\begin{tabular}{|c|l|}
		\hline
		\textbf{Symbol} & \multicolumn{1}{l}{\textbf{Description}} \\ \hline
		$\epsilon$               & Similarity threshold                                 \\ \hline
		$F$               & WiFi fingerprint                                \\ \hline
		$R$               & RSS (average) in dBm                          \\ \hline
		$p$               & Distinct MAC address count                            \\ \hline
		$D$               & Cosine similarity distance                              \\ \hline
		$\alpha$, $\beta$               & Scan result from a list of scan results                              \\ \hline
		$Y$               & Dot product of two WiFi fingerprints                              \\ \hline
		$\omega$               & Number of common mac addresses                              \\ \hline
		$C$               & Cosine similarity score                              \\ \hline
		
	\end{tabular}
\end{table}

In following subsections, we present the details of the WiFi fingerprint clustering and the similarity metrics used in indoor POI extraction process.

\subsection{Unsupervised Indoor POI Extraction}
Research done in \cite{7917558} has experimentally evaluated different clustering algorithms, and has chosen DBSCAN \cite{ester1996density} as the most suitable method because of its ability to form arbitrary shaped clusters. For indoor POI extraction in this paper, we introduce a modified DBSCAN algorithm to cluster the WiFi RSS measurements (i.e. clustered first using GPS data, as mentioned in Section \ref{sec:cloud}). We employ cosine similarity score between two RSS values as the distance metric of the modified DBSCAN algorithm. A cluster (POI) is formed when a user stays for at least $20$ minutes in the same place. Therefore, we choose DBSCAN parameters namely, minimum required points to form a cluster ($minPts$) as $4$ (based on $5$ minute WiFi scan interval), and the cosine similarity threshold ($\epsilon$) to be adaptive, based on the Algorithm \ref{algo:threshold_calculation} (parameter selection is done by experimental evaluation as presented in Table \ref{tab:simThreshold}). Algorithm \ref{algo:POI} explains the procedure of clustering for a given set of WiFi data ($S$), the similarity threshold $\epsilon$, and the minimum points to form a cluster $minPts$. $P$ is the list of output cluster points.

\begin{algorithm}[]
	\SetAlgoLined
	\KwIn{similarity threshold ($\epsilon$), $minPts$, WiFi list ($S$)} 
	\KwOut{Cluster point list ($P$)}
	Visited points ($ V_p$), index $(z_1)$, $P = 0$ \\
	\While{size of $S$ $\ge$ $z_1$}{
		$\alpha =S[z_i]$\\
		\eIf{$\alpha \not\subset V_p$}{
			add $\alpha$ to $V_p$\\
			$N$ = get neighbours of $\alpha$\\
			\If{size of $N \geq minPts$}
			{$z_2=0$\\
				\While{size of $N \geq z_2$}
				{$\beta = N[z_2]$\\
					\eIf{$\beta \not\subset V_p$}
					{add $\beta$ to $V_p$\\
						$Q$ = get neighbours of $\beta$\\
						\If{size of $Q \geq minPts$}{merge $Q$ with $N$ }{}
					}
					{$z_2 = z_2+1$}
				}
				add $N$ to $P$
				
			}
			{} 
		}{
		$z_1 = z_1+1$\;
	}
}
\caption{POI extraction from raw WiFi data}
\label{algo:POI}
\end{algorithm}

The process of obtaining the neighbour points is shown in Algorithm \ref{algo:neighbours}, where inputs are $\alpha$ and $S$, and the output is $N$, which are mentioned in Algorithm \ref{algo:POI}.  The details of the similarity metric is explained in Section \ref{sim_metric}. The computation is done separately for each user. The worst case run time complexity for DBSCAN algorithm is $O(n^2)$, where $n$ is the number of WiFi scan results $(S)$ for a given user.

\begin{algorithm}[!htb]
	\SetAlgoLined
	\KwIn{Scan result ($\alpha$),WiFi list ($S$)}
	\KwOut{Neighbour points ($N$)}
	$N = 0$\\
	\For{every index i in S}
	{$D = $ calculate similarity($\alpha,S[i]$)\\
		$\epsilon$ = calculate threshold($\alpha, S[i]$)\\
		\If{$D \geq \epsilon$}{add $S[i]$ to $N$}
	}
	
	\caption{Obtaining the neighbour points}
	\label{algo:neighbours}
\end{algorithm}

\begin{algorithm}[!htb]
	\SetAlgoLined
	\KwIn{Fingerprints ($F_1$,$F_2$)}
	\KwOut{Similarity threshold ($\epsilon$)}
	\eIf{$F_1 \leq A_L$ and $F_2 \leq A_L$ }{ $\epsilon$ = $\epsilon_L$}	{$\epsilon$ = $\epsilon_H$}
	
	\caption{The process of threshold calculation}
	\label{algo:threshold_calculation}
\end{algorithm}
After getting the final list of clusters, a fingerprint for each cluster (indoor POI) is generated with a unique POI ID (i.e. indoor POI ID is unique to a given GPS stay point). The POI fingerprint ($F$) is denoted as shown in the Equation \ref{eq_fingerprint}, where $M$ is the MAC address, $R$ is the corresponding average RSS in $dBm$, and $p$ is the number of distinct MAC addresses scanned at that POI.

\begin{equation}
\label{eq_fingerprint}
F = {\{M_1,R_1\},\{M_2,R_2\},...,\{M_p,R_p\}}
\end{equation}

\subsubsection{Cosine Similarity}
\label{sim_metric}
We employ cosine similarity as the distance metric in DBSCAN algorithm. The similarity score between two WiFi fingerprints $F_1$ and $F_2$ is calculated as shown below.
\begin{equation}
\label{eq_f1}
F_1 = {\{M^1_1,R^1_1\},\{M^1_2,R^1_2\},...,\{M^1_u,R^1_u\}}
\end{equation}
\begin{equation}
\label{eq_f2}
F_2 = {\{M^2_1,R^2_1\},\{M^2_2,R^2_2\},...,\{M^2_v,R^2_v\}}
\end{equation}
where $u$ and $v$ denote the number of distinct MAC addresses in $F_1$ and $F_2$ respectively. The dot product of RSS in common MAC addresses for the two fingerprints ($Y$) is calculated according to the Equation \ref{eq_dotcommon}, where $w$ denotes the number of common MAC addresses.
\begin{equation}
\label{eq_dotcommon}
Y = \sum_{i=1}^{w} [R^1_i \cdot R^2_i]
\end{equation}
The dot products of each RSS in $F_1$ and $F_2$ are calculated according to the Equations \ref{eq_d1} and \ref{eq_d2} respectively.
\begin{equation}
\label{eq_d1}
d_1 = \sum_{j=1}^{u} [R^1_j \cdot R^1_j]
\end{equation}
\begin{equation}
\label{eq_d2}
d_2 = \sum_{k=1}^{v} [R^2_k \cdot R^2_k]
\end{equation}
The cosine similarity ($C$) between the two WiFi fingerprints is calculated according to the Equation \ref{eq_cosine}.
\begin{equation}
\label{eq_cosine}
C = Y/ (\sqrt{d_1} \times \sqrt{d_2})  \text{ ; where } 0 \leq C \leq 1
\end{equation}

\subsubsection{Impact of Cosine Similarity Threshold}
\label{sec:similarity_evaluation}
We evaluated the different cosine similarity threshold values and their impact on indoor POI extraction. Table \ref{tab:simThreshold} shows the performance of two different similarity threshold values (i.e. adaptive vs. fixed), which are evaluated together with the ground truth labels for the user C in Table \ref{tab_smartphones}. When the threshold is adaptive, the indoor POI extraction result aligns with the ground truth. POI ID $05$ is identified as an additional POI when the threshold value is $0.5$ as highlighted in red in the Table \ref{tab:simThreshold}. When the threshold value is fixed, different POI ID occurs in home environment. This is due to smaller size in the WiFi AP ($<A_L=35 $), observed in that environment. Therefore, we can observe that when the similarity threshold is fixed, even the changes in the size of the scanned AP list (e.g. residential AP list sizes are substantially low, when compared to shopping mall or office AP sizes) have an impact on the WiFi cluster formation.  
\begin{table}[!htb]
	\centering
	\caption{Impact of cosine similarity threshold for indoor POI extraction}
	\label{tab:simThreshold}
	\begin{tabular}{|c|c|c|c|c|}
		\hline
		&                                                                                         &                                                                                       & \multicolumn{2}{c|}{\textbf{POI ID}}                  \\ \cline{4-5} 
		\multirow{-2}{*}{\textbf{Ground Truth}} & \multirow{-2}{*}{\textbf{\begin{tabular}[c]{@{}c@{}}Start Time\\ (HH:mm)\end{tabular}}} & \multirow{-2}{*}{\textbf{\begin{tabular}[c]{@{}c@{}}End Time\\ (HH:mm)\end{tabular}}} & \textit{$\epsilon=$ adaptive} & \multicolumn{1}{c|}{\textit{$\epsilon=0.5$}} \\ \hline
		Home                                    & 00:00                                                                                   & 09:23                                                                                 & 01                & 01                                \\ \hline
		Office                                  & 09:59                                                                                   & 11:34                                                                                 & 02                & 02                                \\ \hline
		Meeting Room                            & 11:58                                                                                   & 14:57                                                                                 & 03                & 03                                \\ \hline
		Canteen                                 & 15:29                                                                                   & 16:39                                                                                 & 04                & 04                                \\ \hline
		Office                                  & 16:44                                                                                   & 17:09                                                                                 & 02                & 02                                \\ \hline
		Home                                    & 17:49                                                                                   & 18:54                                                                                 & 01                & 01                                \\ \hline
		Home                                    & 18:59                                                                                   & 20:39                                                                                 & 01                & \cellcolor[HTML]{FD6864}05        \\ \hline
		Home                                    & 20:48                                                                                   & 23:53                                                                                 & 01                & 01                                \\ \hline
	\end{tabular}
\end{table}

\subsubsection{Popular POI Among Users}
\label{sec:commonPOI}
Knowing the popular POI among users is as equally important as knowing individual indoor POI when conducting user mobility analysis. We make use of Louvain method for community detection \cite{blondel2008fast} to gain insights on popular POI among users. In a given indoor environment, let the number of POI be $\lambda$, and the number of pair-wise cosine similarity scores ($I$) is calculated according to the Equation \ref{eq_pairwise}. 
\begin{equation}
\label{eq_pairwise}
I = \frac{h!}{2!(h-2)!}\quad
\end{equation}
The Louvain algorithm takes $I$ as the input and obtains the optimum partitioning among POI (nodes) by comparing pair-wise similarity (edges) scores, and provides the modularity as the output. 
The results of indoor POI extraction are presented in the following subsection.
\subsection{Results}

We collected WiFi and GPS data from a set of users (who use different smartphone models) together with the ground truth labels of the POI they visited. The experimental results for single user and multi user indoor POI extraction are presented in the following subsections. 
\subsubsection{Single User POI}

For single user POI identification, we did an experiment to identify the POI when a single user visits the same POI multiple times. The WiFi clustering results are compared along with the ground truth. Table \ref{tab_individualiuser} presents the single user POI identification for user H, during one year time duration of POI visits in Changi General Hospital. According to Table \ref{tab_individualiuser}, the POI ID is different for different locations (i.e. WiFi clusters) inside the building. The proposed clustering technique is capable to detect when the user revisits POI ID $04$, $07$, $10$, $11$ and $12$.

\begin{table}[!htb]
	\centering
	\caption{Single user indoor POI results}
	\label{tab_individualiuser}
	\begin{tabular}{|c|c|c|c|c|}
		\hline
		\textbf{\begin{tabular}[c]{@{}c@{}}Ground\\ Truth\end{tabular}}                  & \textbf{\begin{tabular}[c]{@{}c@{}}Date\\ (yyyy-mm-dd)\end{tabular}} & \textbf{\begin{tabular}[c]{@{}c@{}}Start Time\\ (HH:mm)\end{tabular}} & \textbf{\begin{tabular}[c]{@{}c@{}}End Time\\ (HH:mm)\end{tabular}} & \textbf{\begin{tabular}[c]{@{}c@{}}POI\\ ID\end{tabular}} \\ \hline
		\multirow{3}{*}{\begin{tabular}[c]{@{}c@{}}Center for\\ Innovation\end{tabular}} & 2019-01-22                                                           & 11:32                                                                 & 11:57                                                               & \multirow{3}{*}{04}                                       \\ \cline{2-4}
		& 2019-02-21                                                           & 16:09                                                                 & 17:36                                                               &                                                           \\ \cline{2-4}
		& 2019-08-06                                                           & 10:02                                                                 & 10:52                                                               &                                                           \\ \hline
		\begin{tabular}[c]{@{}c@{}}Level 6\\ Room\end{tabular}                           & 2019-04-02                                                           & 10:58                                                                 & 12:23                                                               & 06                                                        \\ \hline
		\multirow{2}{*}{\begin{tabular}[c]{@{}c@{}}Main Board\\ Room\end{tabular}}       & 2019-05-08                                                           & 16:58                                                                 & 18:48                                                               & \multirow{2}{*}{07}                                       \\ \cline{2-4}
		& 2019-08-13                                                           & 15:10                                                                 & 15:35                                                               &                                                           \\ \hline
		\multirow{2}{*}{\begin{tabular}[c]{@{}c@{}}Level 8\\ Room\end{tabular}}          & 2019-07-11                                                           & 09:44                                                                 & 10:24                                                               & \multirow{2}{*}{10}                                       \\ \cline{2-4}
		& 2019-07-18                                                           & 08:39                                                                 & 09:32                                                               &                                                           \\ \hline
		\multirow{3}{*}{\begin{tabular}[c]{@{}c@{}}Level 7\\ Room\end{tabular}}          & 2019-07-24                                                           & 15:14                                                                 & 16:14                                                               & \multirow{3}{*}{11}                                       \\ \cline{2-4}
		& 2019-07-31                                                           & 08:48                                                                 & 09:59                                                               &                                                           \\ \cline{2-4}
		& 2019-08-01                                                           & 11:57                                                                 & 14:14                                                               &                                                           \\ \hline
		\multirow{2}{*}{Ward 45}                                                         & 2019-08-02                                                           & 13:12                                                                 & 15:54                                                               & \multirow{2}{*}{12}                                       \\ \cline{2-4}
		& 2019-10-15                                                           & 15:15                                                                 & 15:54                                                               &                                                           \\ \hline
	\end{tabular}
\end{table}

\subsubsection{Multi User POI}

In this subsection, we intend to identify the indoor POI, which are popular among multiple users. We select Changi City Point (CCP), where $11$ users (i.e. A to L in Table \ref{tab_smartphones}) from our experiment visit for the purpose of shopping/dining, for a duration of 3 months. The clustering results detected $41$ indoor POI at the CCP. $820$ pair-wise similarities are given as the input to Louvain algorithm for community detection as shown in the Equation \ref{eq_pairwise}. We evaluated the modularity score for different similarity threshold values as shown in the Table \ref{tab_modularity} to understand the optimum partition for communities. Since, different POI are different in terms of area size (e.g. food court is larger than clothing shop), our objective is to detect even the smallest POI visited by users. Therefore, for POI identification we selected $0.5$ as the partitioning threshold for community detection. 

\begin{table}[htb!]
	\centering
	\caption{Louvain modularity score for different partitioning thresholds}
	\label{tab_modularity}
	\begin{tabular}{|c|c|}
		\hline
		\textbf{Threshold Value} & \textbf{Modularity Score} \\ \hline
		0.2                & 0.625                       \\ \hline
		0.3                & 0.803                       \\ \hline
		0.4                & 0.766                       \\ \hline
		0.5                & 0.692                       \\ \hline
	\end{tabular}
\end{table}

\begin{figure}[H] 
	\centering
	\includegraphics[width=0.49\textwidth]{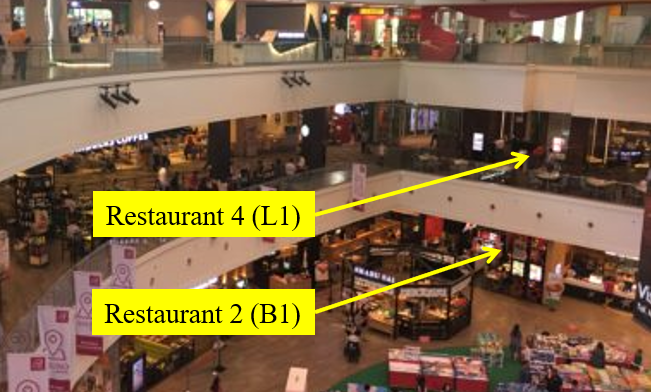} 
	\caption{Changi City Point Basement 1 and Level 1 Layout.}
	\label{fig:layout_ccp} 
\end{figure}

Table \ref{tab:multiuser} shows the details common POI visited by the $11$ users in Changi City Point. The shopping mall is a three-storey building with Basement 1 (B1), Level 1 (L1), and Level 2 (L2). Users from the study carried on with their normal routine to the mall for shopping/dining purposes. From the table we can observe that, users H, J, and K visited $3$ different restaurants (at 3 different timing) in B1, denoted with indoor POI IDs $00$, $11$, and $29$ respectively. The Restaurant 4 in L1 also obtained the same indoor POI ID (i.e. POI ID $11$) as the Restaurant 2 in B1, where user J visited. When we checked the shopping mall layout, we observed that even those two restaurants are in two different levels, they are located right above one another, as shown in Figure \ref{fig:layout_ccp}. Also, there is a wide opening between them, which leads to similar WiFi measurement at those two places. Another observation is that, a large area like food court (almost half the size of L2) is divided into multiple POI, since users sat on various places and the WiFi RSS measurement is fluctuating due to large crowd.

\begin{table}[htb!]
	\centering
	\caption{Common POI among different users in Changi City Point shopping mall}
	\label{tab:multiuser}
	\begin{tabular}{|c|c|c|c|}
		\hline
		\textbf{\begin{tabular}[c]{@{}c@{}}Floor Level\end{tabular}} & \textbf{\begin{tabular}[c]{@{}c@{}}Ground Truth\end{tabular}} & \textbf{\begin{tabular}[c]{@{}c@{}}POI ID\end{tabular}} & \textbf{User(s)} \\ \hline
		\multirow{5}{*}{B1}                                            & Restaurant 1                                                     & 00                                                         & H                \\ \cline{2-4} 
		& Restaurant 2                                                     & \cellcolor[HTML]{FD6864}11                                                         & J                \\ \cline{2-4} 
		& Restaurant 3                                                     & 29                                                         & K                \\ \cline{2-4} 
		& Drink Shop                                                       & 05                                                         & G                \\ \cline{2-4} 
		& Utility Store                                                    & 13                                                         & K                \\ \hline
		\multirow{2}{*}{L1}                                            & Restaurant 4                                                     & \cellcolor[HTML]{FD6864}11                                                         & B,H              \\ \cline{2-4} 
		& Clothing Shop 1                                                  & 27                                                         & A                \\ \hline
		\multirow{8}{*}{L2}                                            & Clothing Shop 2                                                  & 08                                                         & C                \\ \cline{2-4} 
		& Clothing Shop 3                                                  & 09                                                         & C,D,I            \\ \cline{2-4} 
		& \multirow{6}{*}{Food Court}                                      & \cellcolor[HTML]{FD6864}01, 02                                                      & A                \\ \cline{3-4} 
		&                                                                  &\cellcolor[HTML]{FD6864}03, 05                                                     & C                \\ \cline{3-4} 
		&                                                                  & \cellcolor[HTML]{FD6864}07, 10                                                     & D                \\ \cline{3-4} 
		&                                                                  & \cellcolor[HTML]{FD6864}12, 15                                                     & F                \\ \cline{3-4} 
		&                                                                  & \cellcolor[HTML]{FD6864}21, 23                                                     & J                \\ \cline{3-4} 
		&                                                                  & \cellcolor[HTML]{FD6864}24, 25                                                     & K                \\ \hline
	\end{tabular}
\end{table}

\section{Neighborhood Activity Study}
\label{sec_4}

In an urban area, majority of the residents tend to visit nearby places of home or office for shopping and leisure activities during their free time. Especially in dense areas, where high rise buildings are common as residences, conventional GPS clustering approach using GPS data may indicate such a building as one POI, but in reality there are many possible POIs (e.g. convenient store, common area, BBQ pit etc.) in a multi-storey setting.
This is due to the dimensionality nature of GPS data, and GPS data alone cannot provide accurate information on stay points at micro level. Moreover, it is useful to understand the user stay points in the residential(home) neighborhood.
We define such stay points or places of short duration as neighborhood activity, and exploit WiFi fingerprint along with GPS data to identify such neighborhood activity.

\subsection{Neighborhood Activity Data Processing Architecture}

\begin{figure}[H] 
	\centering
	\includegraphics[width=0.49\textwidth]{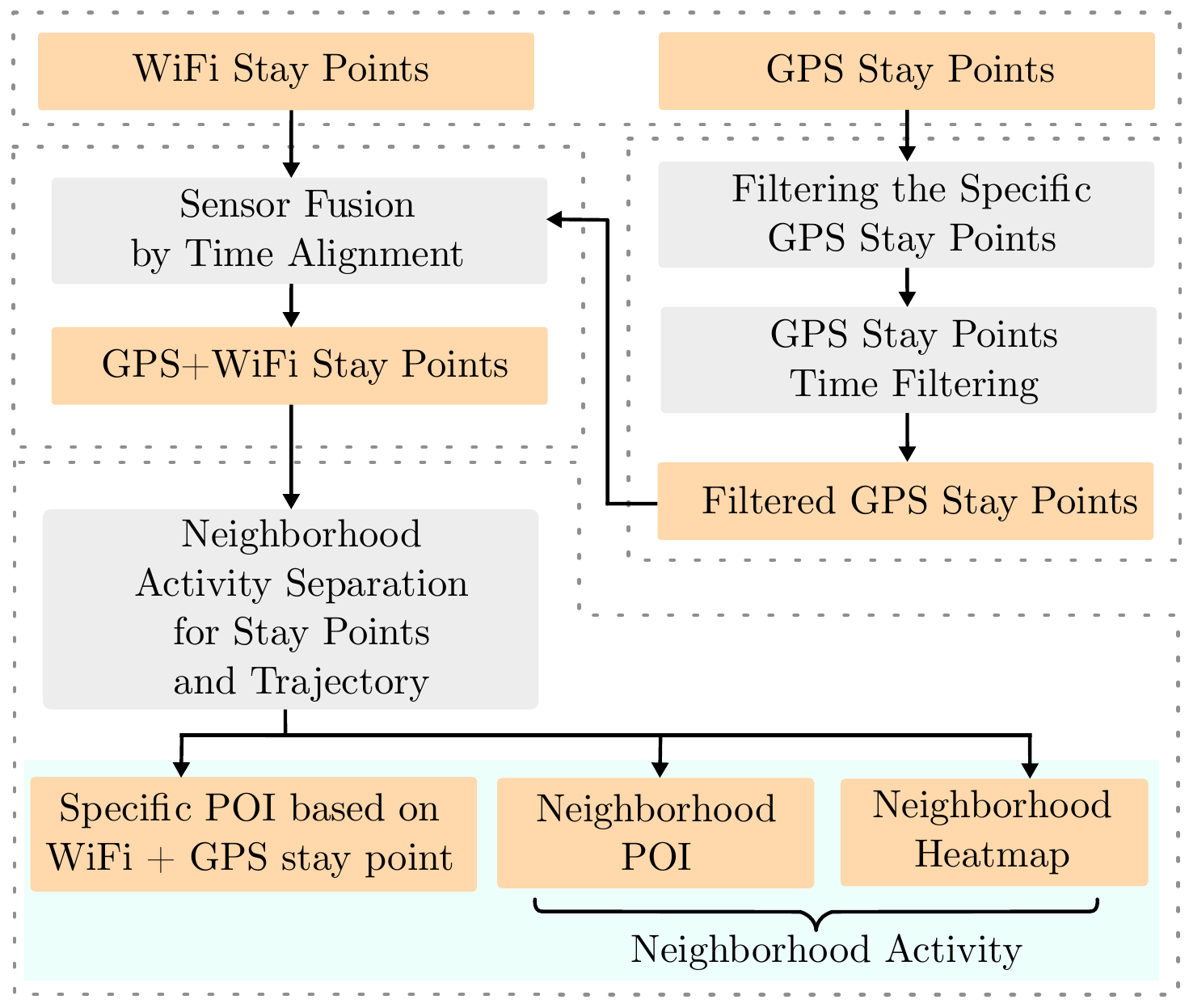} 
	\caption{Neighborhood activity data processing pipeline.}
	\label{fig:dataflow_neighborhood} 
\end{figure}

In order to extract the neighborhood activity from the trajectory, we leverage the concept of sensor fusion to combine GPS and WiFi information sources. 
The overall process of the neighborhood activity extraction is illustrated in Figure~\ref{fig:dataflow_neighborhood}. 

There are two main data sources used in the processing stage, which are WiFi and GPS stay points. First, we identify the GPS stay points and label them accordingly to understand the characteristics of each GPS POI. Subsequently, we filter it by time to study particular point of interest. Note that while we use a user’s house as point of interest, it could also be an office or any other GPS POI. The filtered GPS stay points will be fused with WiFi stay points to generate GPS+WiFi stay points (WiFi stay points are generated according to Algorithm \ref{algo:POI}). Among all the GPS and WiFi stay points identified based on the duration of stay, one can easily deduce home or office heuristically, which both are stay points with the longest stay durations. The rest of the GPS and WiFI stay points will be the neighborhood POI. The remaining raw GPS points (moving points) that occurred between neighborhood POI and a specific POI can be further converted into heat maps to capture potential neighborhood activity that does not form a stay point.

\begin{figure}[H] 
	\centering
	\begin{subfigure}[t]{0.3\textwidth}
		\centering
		\includegraphics[width=1.0\textwidth]{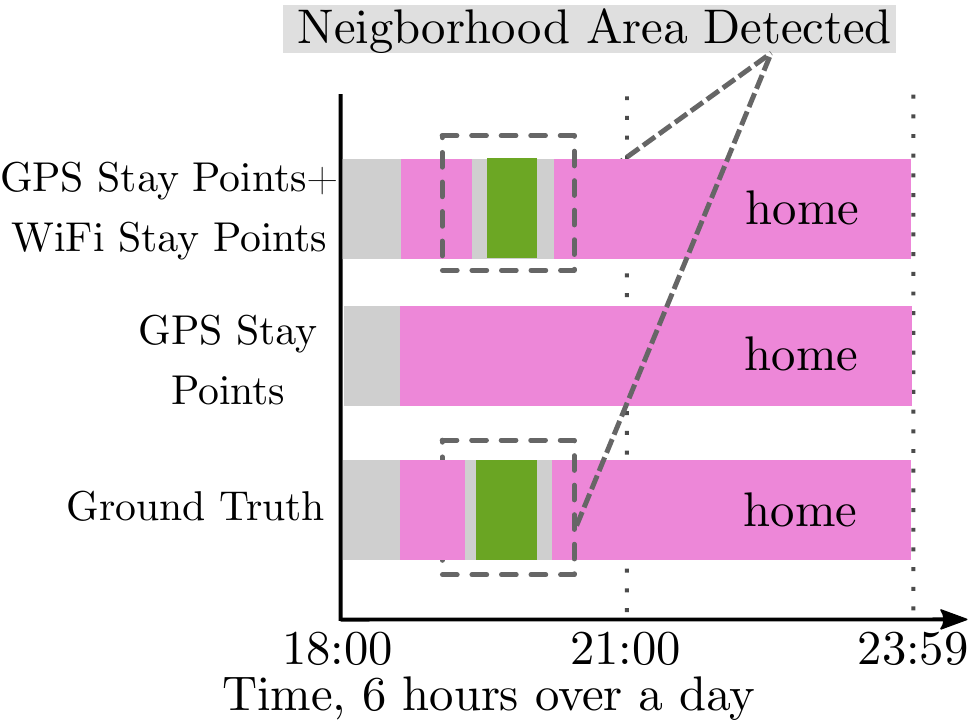} 
		\caption{Stay points extraction methods comparison}
		\label{fig:clus_compare_barchart}
	\end{subfigure}
	~
	\begin{subfigure}[t]{0.17\textwidth}
		\centering
		\includegraphics[width=1.0\textwidth]{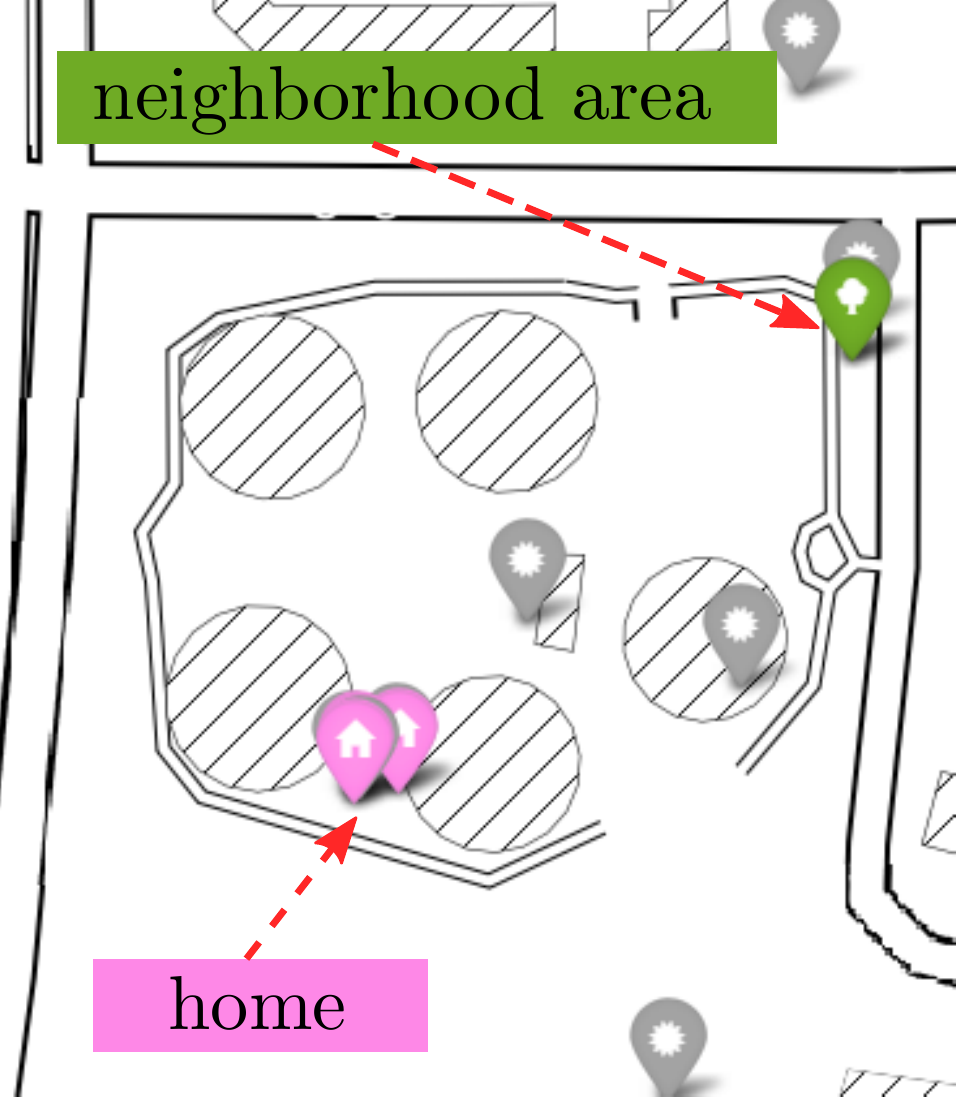} 
		\caption{Neighborhood activity's visualization}
		\label{fig:clus_compare_gps}
	\end{subfigure}
	
	\caption{Toy example of performing neighborhood activity extraction.}
	\label{fig:cluster_comparison}
\end{figure}

As a proof of concept, we perform a simple case study for an user H using WiFi and GPS data for $6$ hours period of the day of 18 September 2019.
We compared the methods between different GPS and GPS+WiFi stay points as illustrated in Figure~\ref{fig:clus_compare_barchart}.
We observe that GPS+WiFi stay points data fusion method is able to detect the neighborhood activity, where the GPS stay points method is not capable.
Subsequently, visualization of the stay points are shown in Figure~\ref{fig:clus_compare_gps}, where gray icons represents traveling GPS data, and pink and green icon denotes home and neighborhood POI.
The neighborhood area is located not far away from the residential area, which is less than $100m$.
Using only GPS stay points, it may appear that neighborhood activity is almost similar location to stay points, and thus clustering as same stay points.
Therefore, we are able to detect the neighborhood activity accurately, using GPS+WiFi data sources, compared to using GPS only as data source.

\subsection{Results}
The results of the neighborhood activity study are presented in the following subsections. 

\subsubsection{Single User Neighborhood Activity}

\begin{figure*}[!h] 
	\centering
	\begin{subfigure}{0.32\textwidth}
		\includegraphics[width=\linewidth]{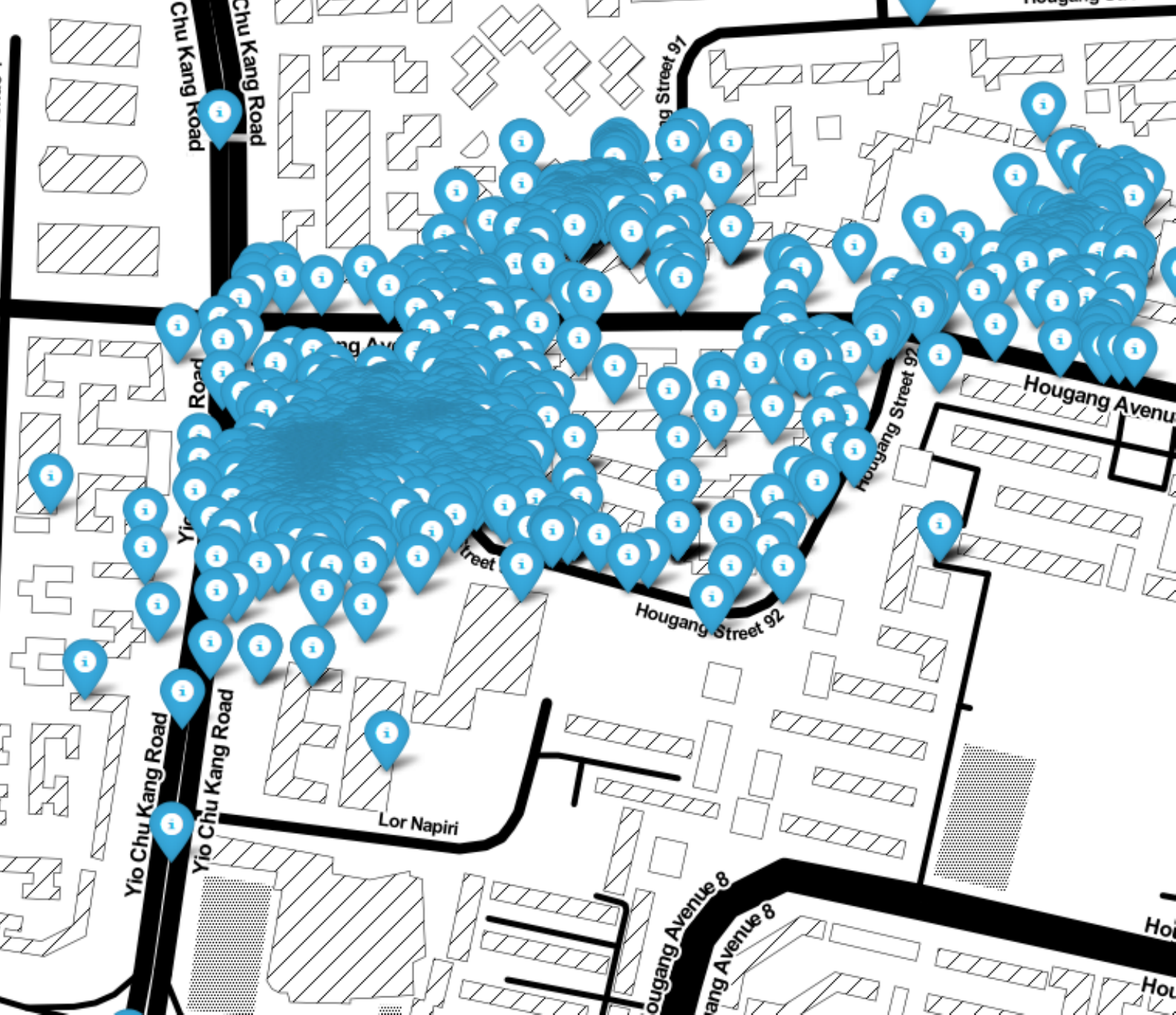}
		\caption{Raw GPS Data} \label{fig:neighbourhood_comparison_raw}
	\end{subfigure}
	\begin{subfigure}{0.32\textwidth}
		\includegraphics[width=\linewidth]{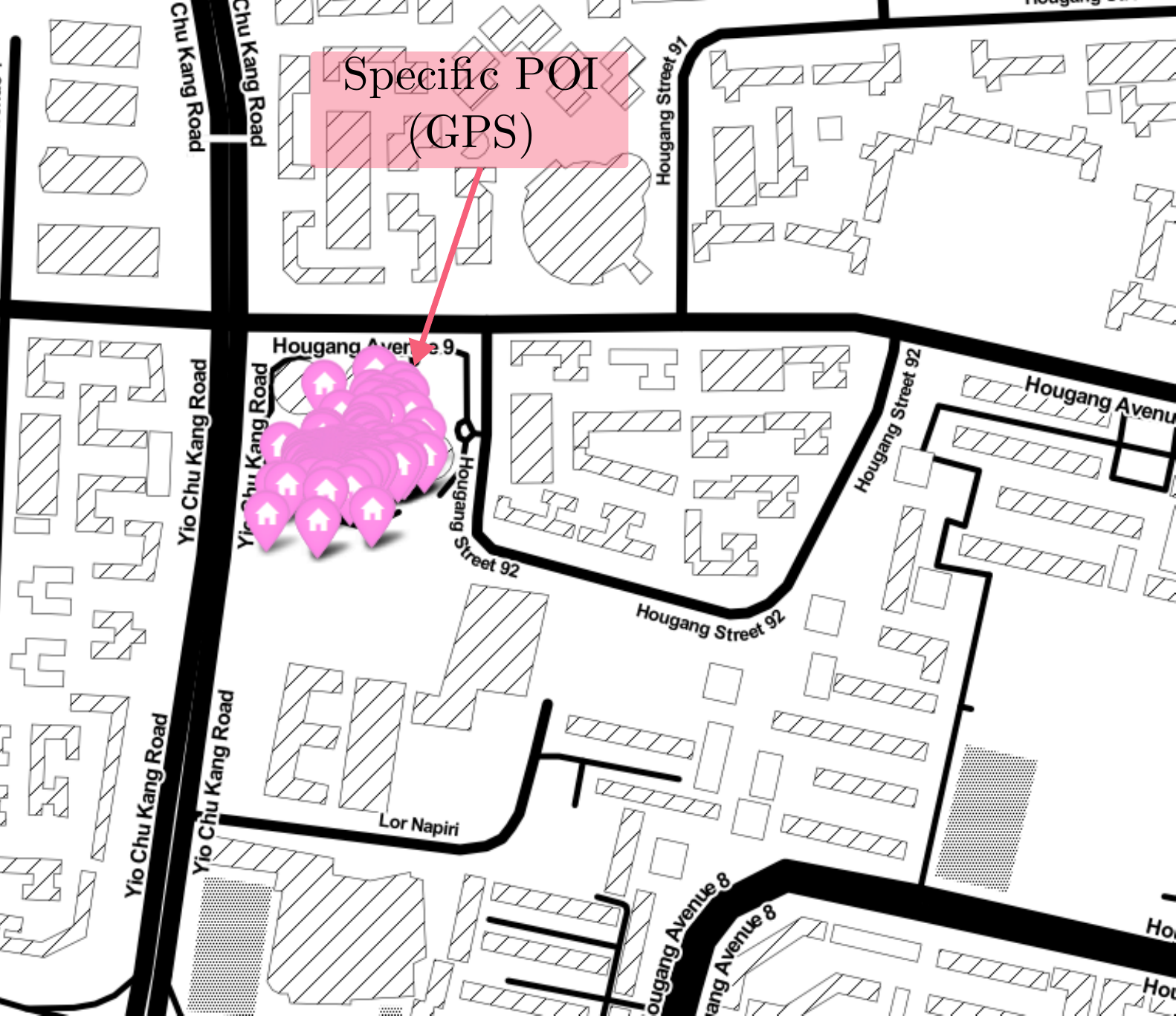}
		\caption{GPS Stay Points} \label{fig:neighbourhood_comparison_gps}
	\end{subfigure}
	\begin{subfigure}{0.32\textwidth}
		\includegraphics[width=\linewidth]{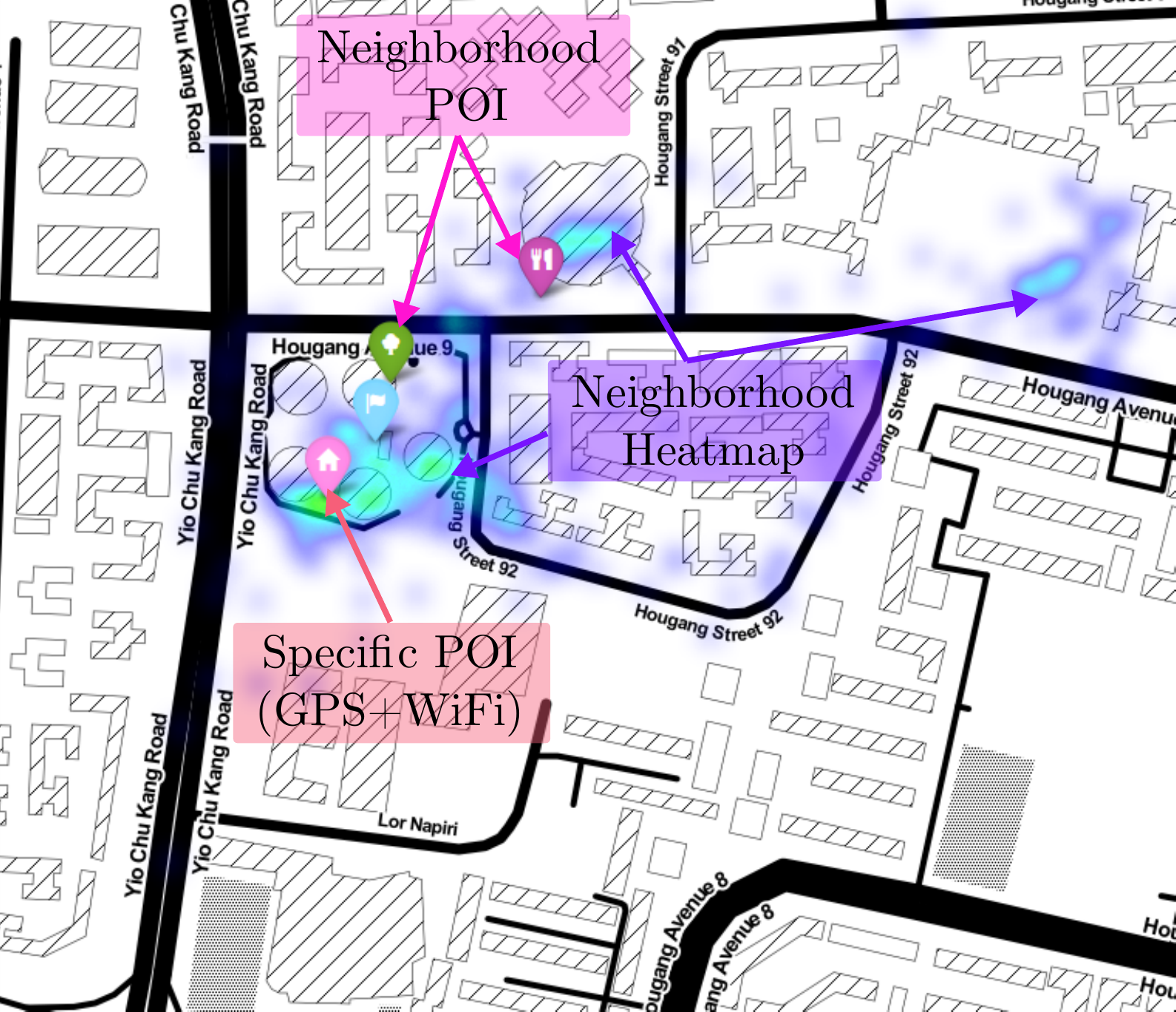}
		\caption{Neighborhood Activity by GPS+WiFi Stay Points} \label{fig:neighbourhood_comparison_wifi}
	\end{subfigure}
	\caption{Comparison of raw data, GPS stay points, and neighborhood activity by GPS+WiFi stay points, for user H, from 01 April 2019 to 01 December 2019.}
	\label{fig:neighbourhood_comparison} 
\end{figure*}

Using the aforementioned extraction techniques, we perform study on the user H over 8 months starting from 01 April 2019 to 01 December 2019.
The raw GPS data is extracted based on the home's location of user H, where unrelated GPS data are filtered in order to help us focus on that particular region.
Note that, same temporal notion of the home stay point is applied to WiFi data to study neighborhood activity.
Figure~\ref{fig:neighbourhood_comparison} shows the comparison of neighborhood activity obtained by GPS and GPS+WiFi.
From Figure~\ref{fig:neighbourhood_comparison_raw}, we can observe from the raw data that user H has traveled to nearby places from home, where GPS stay point in Figure~\ref{fig:neighbourhood_comparison_gps} fails to detect such events. 
It could happen because of the POI user traveled is a nearby location, which is indistinguishable by the GPS data.
Using the GPS+WiFi stay points in Figure~\ref{fig:neighbourhood_comparison_wifi}, we are able to detect neighborhood POI that the an user has visited (green, blue, and purple icons).
Also, since WiFi data is fused with GPS stay points, specific POI location can be exactly identified, and the remaining moving raw GPS points are converted to heat map.
The blue and green icons represent the housing recreational facilities, while the purple icon is referred to a nearby community mall.
From the heat map, we notice some hot spots from the heat maps that user has visited while at the home region, but it does not form a stay point.
To check whether that particular hot spots observed from the heat maps, we also validate the corresponding location with user H with each neighborhood activity shown in the blue patch on the top right corner.
It turns out that the user H only visit the location for a short period of time, which stay duration is lesser than the predefined stay time threshold.
Hence, stay point is not formed due to short duration, and only can be observed through heat maps.
In a nutshell, we have demonstrated that through combination of GPS along with WiFi stay points, neighborhood activity can be detected to provide in-depth information to daily trajectory of the user.

To contrast, there is no traveling event around the neighborhood captured using GPS stay points detection as shown in Figure~\ref{fig:clus_compare_barchart} and Figure~\ref{fig:neighbourhood_comparison_gps}. Therefore, we have demonstrated using GPS+WiFi data, it is possible to detect neighborhood activity within a region to further enhance user's trajectory data context.

\subsubsection{Multi User Neighborhood Activity}

\begin{figure*}[!h] 
	\centering
	\begin{subfigure}{0.32\textwidth}
		\includegraphics[width=\linewidth]{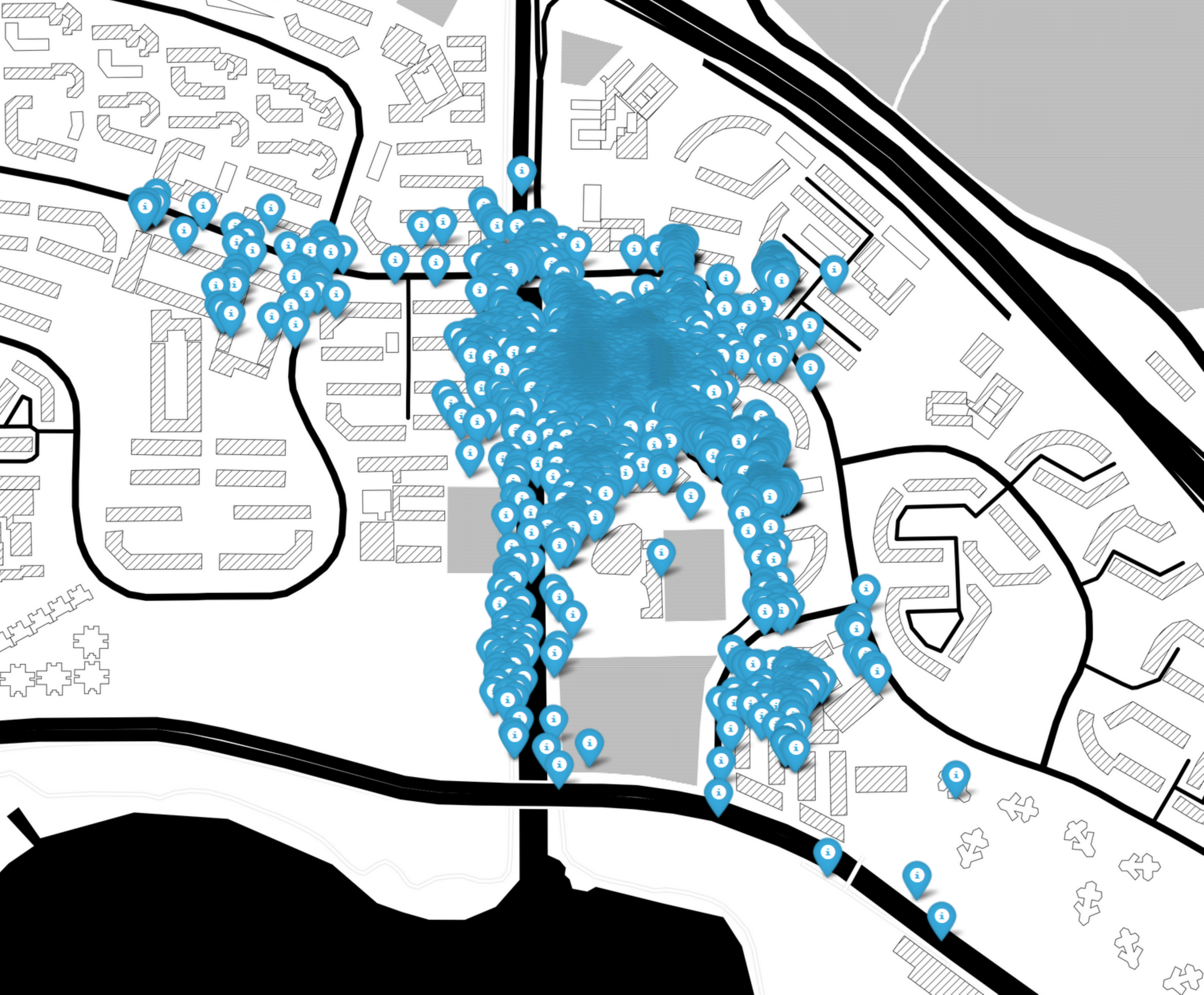}
		\caption{Raw GPS Data} \label{fig:neighbourhood_multi_raw}
	\end{subfigure}
	\centering
	\begin{subfigure}{0.32\textwidth}
		\includegraphics[width=\linewidth]{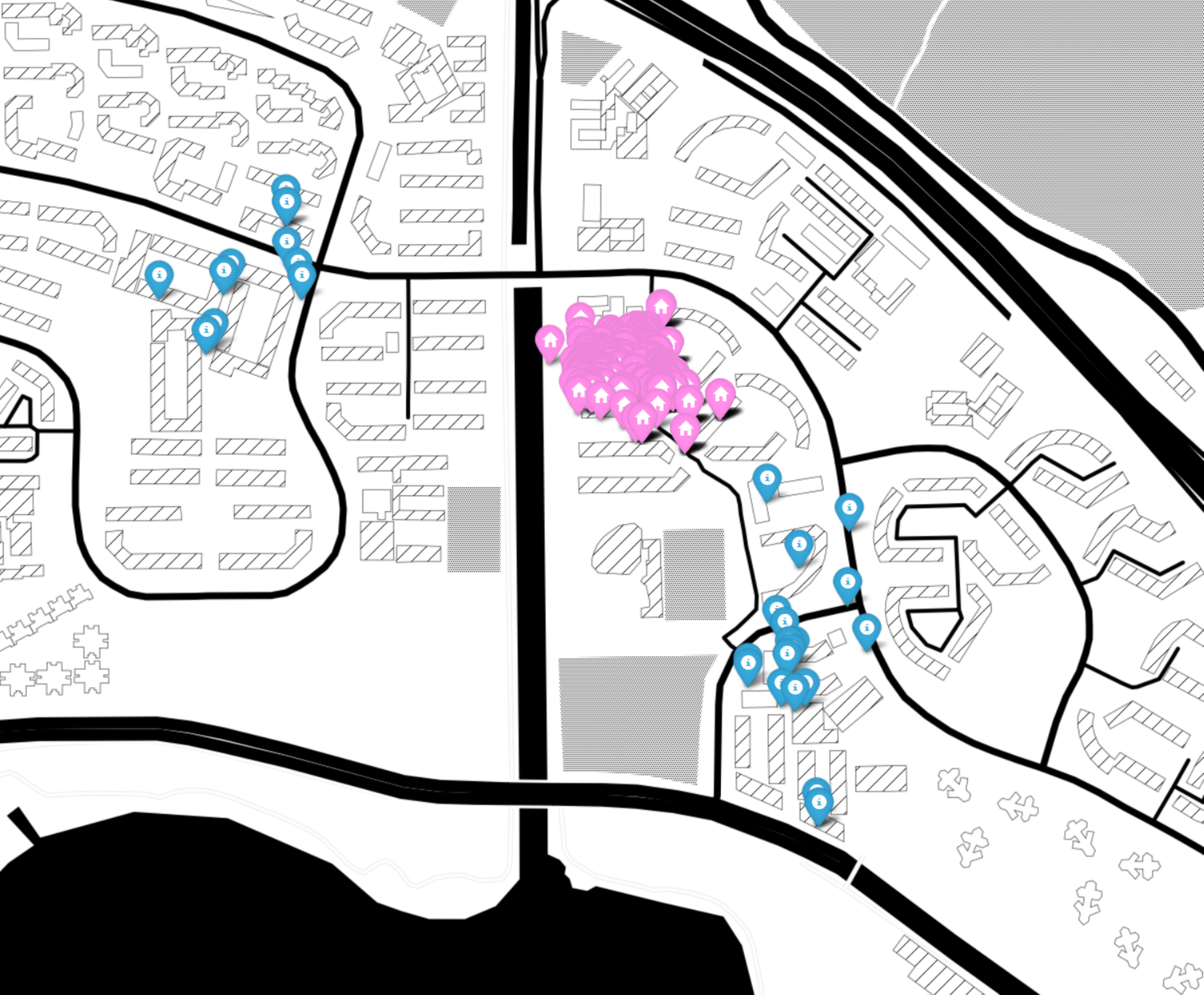}
		\caption{GPS Stay Points by 3 Users} \label{fig:neighbourhood_multi_gps}
	\end{subfigure}
	\begin{subfigure}{0.32\textwidth}
		\includegraphics[width=\linewidth]{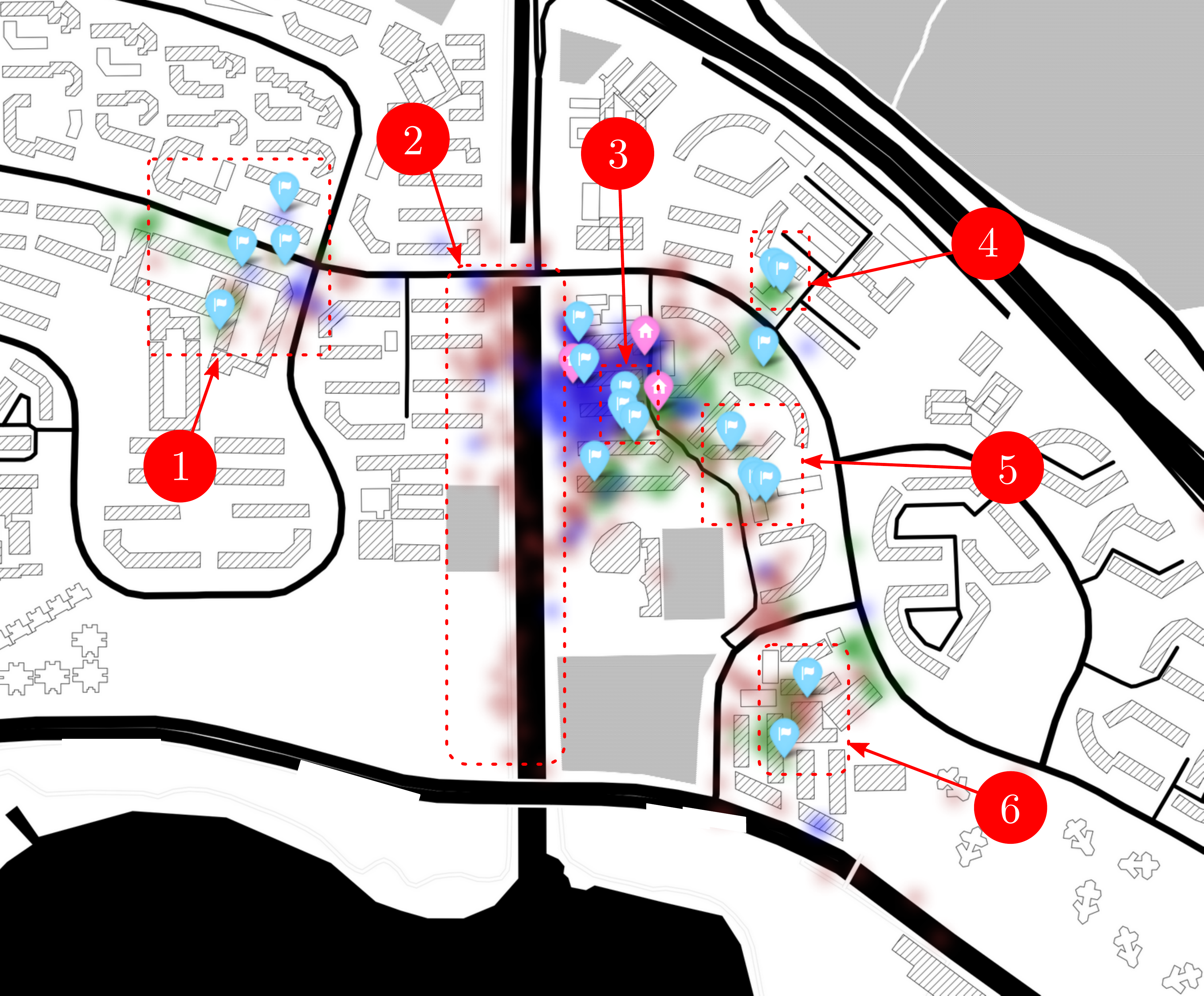}
		\caption{Neighborhood Activity by 3 Users using GPS+WiFi} \label{fig:neighbourhood_multi_wifi}
	\end{subfigure}
	\caption{Comparison of the neighbourhood activity between 3 users, data collected from 01 August 2020 to 15 March 2021. Note that heatmaps in GPS+WiFi is represented by three different colours (red, green, and blue) to indicate different users’ trajectory.}
	\label{fig:neighbourhood_multi} 
\end{figure*}

Figure \ref{fig:neighbourhood_multi} shows the neighbourhood activity obtained using the proposed method for three users who reside in the same neighborhood. These users are out of the Table \ref{tab_smartphones} and their POI visit ground truth is unknown. Figure \ref{fig:neighbourhood_multi_raw} shows the raw GPS points for the three users. Figure \ref{fig:neighbourhood_multi_gps} shows the GPS stay points by the three users, while Figure \ref{fig:neighbourhood_multi_wifi} shows the GPS+WiFi stay points (blue pins) and the home locations (pink pins) for each user. By comparing the figures, we can observe that WiFi data cleans up a lot of inaccuracies of the GPS stay points for the three users.

One can notice that in the areas of 1, 5, and 6, the POIs become clearer in Figure \ref{fig:neighbourhood_multi_wifi} as compared to Figure \ref{fig:neighbourhood_multi_gps}. The WiFi information help us to identify GPS stay point that belong to the same POI. The heatmap in area 2, indicate the users walk along the river side, which is missing from Figure \ref{fig:neighbourhood_multi_gps}. In addition, while the users stay in area 3, there are quite a number of POIs in area 3 we well (those believe to be void deck directly underneath of the user's home), and once again, these POIs are not visible by GPS in Figure \ref{fig:neighbourhood_multi_gps}, as they all are identified as user's home. Finally, new POI is identified in area 4. 

\section{Micro Mobility Study}
\label{sec_5}

Users in the same residential neighborhood might share similar mobility patterns around the neighborhood. 
We define such mobility patterns as micro mobility of the neighborhood, and extract the mobility paths through a combination of WiFi and GPS data.
The following subsections present the data processing technique and the results for micro mobility analysis.

\subsection{Micro Mobility Path Extraction}
To further study the mobility data using both WiFi and GPS data, we propose a data processing pipeline as shown in Figure~\ref{fig:dataflow_micromobility} below.

\begin{figure}[!htb] 
	\centering
	\includegraphics[width=0.49\textwidth]{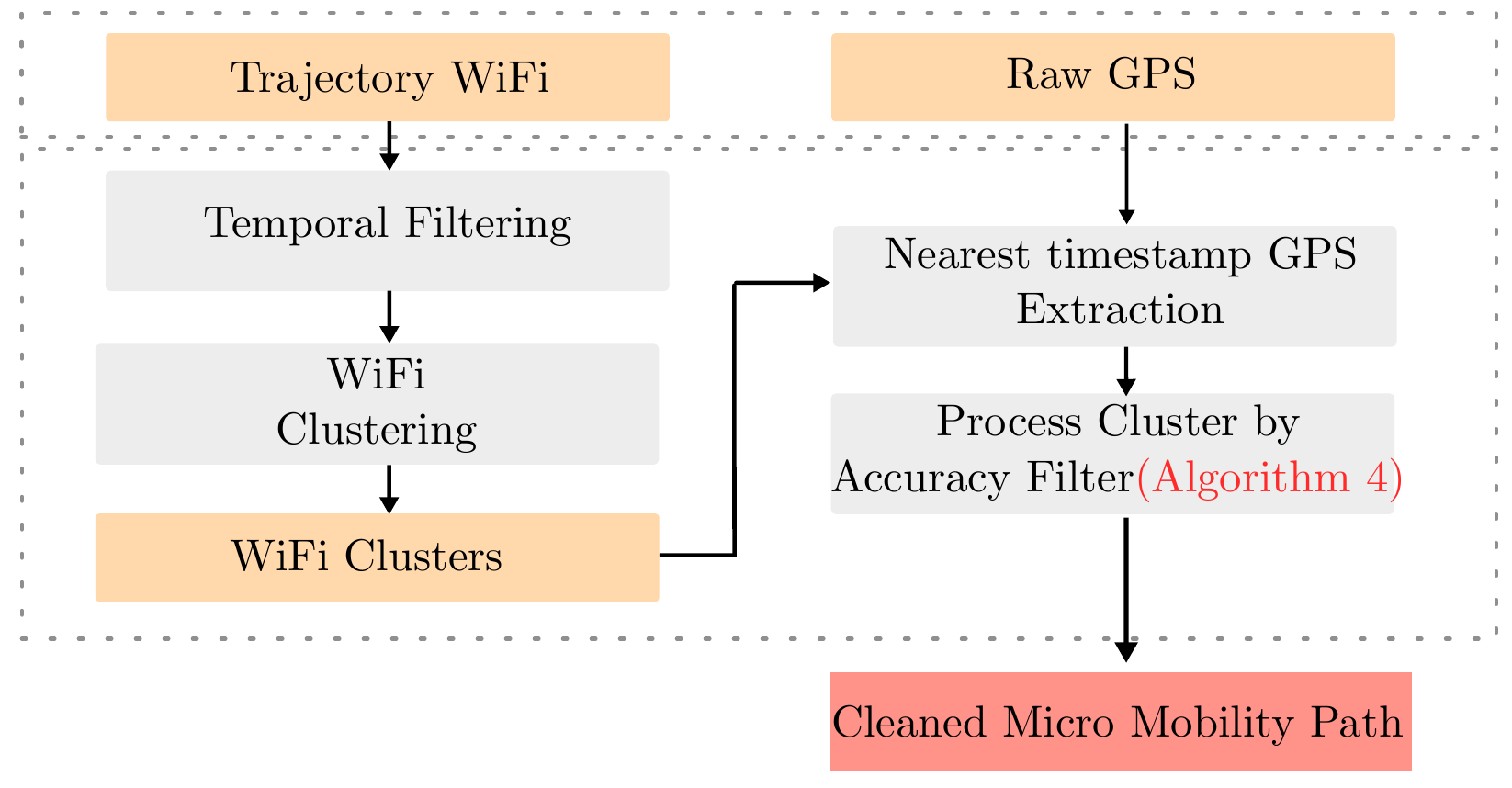} 
	\caption{Micro mobility analysis data processing pipeline} 
	\label{fig:dataflow_micromobility} 
\end{figure}

Using the trajectory data obtained from the GPS stay points, we perform the timeline extraction to obtain the exact moment of WiFi samples needed to further study micro mobility.
Neighborhood WiFi trajectory data are clustered together, using DBSCAN for all users who live in the same neighborhood.
WiFi based clustering process is shown in Algorithm \ref{algo:micromobility}.
\begin{algorithm}[!htb]
	\SetAlgoLined
	\KwIn{Trajectory WiFi ($S_T$) and GPS ($L_T$), $\epsilon$, $minPts$}
	\KwOut{Processed WiFi based clusters ($N_p$)}
	$N_P = 0$\\
	Cluster list ($C$) = DBSCAN ($S_T$, $\epsilon$, $minPts$) \\
	\For{every index i in C}
	{$L_n = $ nearest GPS($t,C[i],L_T$)\\
		\eIf{accuracy $(a) \leq a_L$}{add average $C[i]$ to $N_P$}{get lowest accuracy, add to $N_P$}
	}
	\caption{The process of extracting WiFi based micro mobility clusters}
	\label{algo:micromobility}
\end{algorithm}

Since we want to identify similar trajectory path, our objective in this scenario is different from that of identifying indoor stay points in Section \ref{sec_3}. To understand the micro mobility, we need to identify the travel path, not the stay point. In other words, our objective is to clear up a messy interpretation of GPS map (as shown in Figure \ref{fig:micro_raw}) into a clearer map (as shown in Figure \ref{fig:micro_clustered}) Therefore, DBSCAN parameters are different in this scenario. We set $minPts=1$ as we need to include every WiFi scan result into the clustering process. Since our sampling rate is low ($5$ min), a user can travel a substantial distance during that time period. Therefore, we want to include every scan result to the clustering process, when identifying the mobility pattern. By evaluating the clustering results, we choose threshold level for cluster formation ($\epsilon$) to provide enough number of clusters to represent the user travel path which reducing the average distance error in WiFi based GPS clusters. Moreover, the number of APs we observe in outdoors are below the low AP level ($A_L$).

Once the clustering is completed, we obtain nearest GPS point for each cluster point's timestamp. If there are more than one member in a particular cluster, we obtain the average of the nearest GPS points with high accuracy (i.e. accuracy $\leq a_L=25m$) and represent one WiFi based cluster with one GPS point. If all the members in a cluster indicate low GPS accuracy (i.e. accuracy $> a_L=25m$), we get the lowest accuracy value GPS point (which means the highest GPS accuracy), and discard the rest of the members in the cluster. 

\begin{figure}[!htb] 
	\centering
	\includegraphics[width=0.49\textwidth]{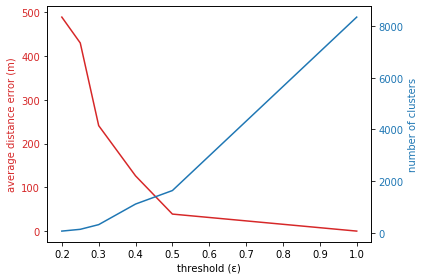} 
	\caption{Comparison of different threshold values with average distance error (m) and number of clusters} 
	\label{fig:thresholds} 
\end{figure}

Figure \ref{fig:thresholds} shows the comparison between the number of clusters, average distance error (in meters) of the cluster points, vs. different threshold values for WiFi based clustering. 

We can observe from the Figure \ref{fig:thresholds}, that when the threshold value increases the number of clusters also increase and the average distance error decreases. Our objective is to reduce the number of clusters (to obtain a clearer mobility path) and to reduce the average distance error. When $\epsilon=0.25$, we get $140$ clusters with $430.1$ meters of distance error. In contrast when $\epsilon=0.3$, we get $321$ clusters with $241.1$ meters of distance error. Therefore, by considering this trade-off we select $\epsilon=0.3$ as the threshold value for WiFi based clustering. It gives enough number of clusters to represent a messy GPS micro mobility path into a clearer path while having a reduced average distance error.
\subsection{Results}

\begin{figure*}[!htb]
	\centering
	\begin{subfigure}{0.49\textwidth}
		\includegraphics[width=\linewidth]{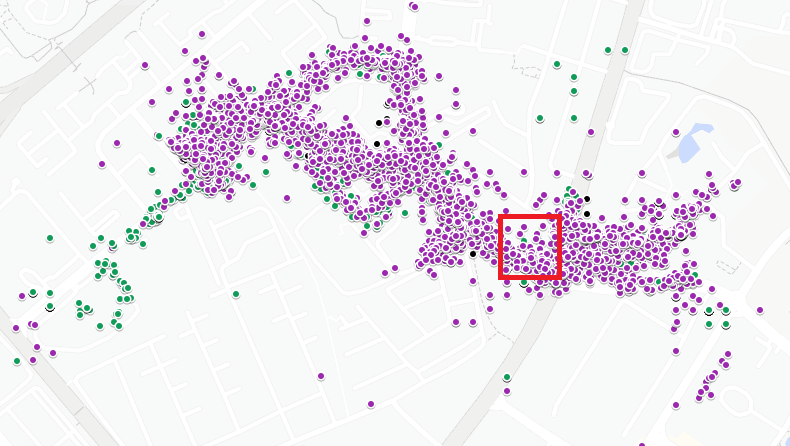}
		\caption{Raw GPS data (before WiFi based clustering), denoted by users \color{purple}{Purple - L}, \color{green}{Green - A}, \color{black}{and Black - B.}} \label{fig:micro_raw}
	\end{subfigure}
	\hspace*{\fill} 
	\begin{subfigure}{0.49\textwidth}
		\includegraphics[width=\linewidth]{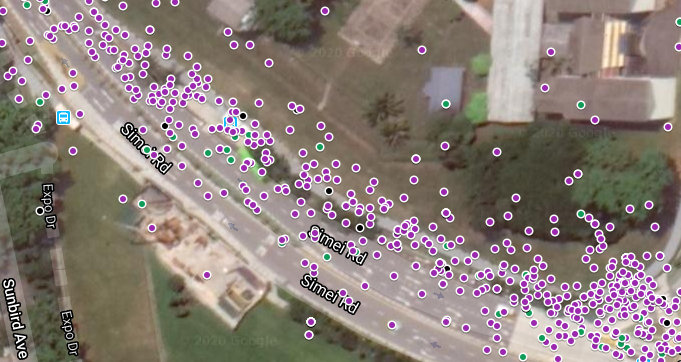}
		\caption{Zoomed in satellite view of raw GPS data fluctuated along a sheltered walkway, denoted by users \color{purple}{Purple - L}, \color{green}{Green - A}, \color{black}{and Black - B.}} \label{fig:satellite_raw}
	\end{subfigure}
	\begin{subfigure}{0.49\textwidth}
		\includegraphics[width=\linewidth]{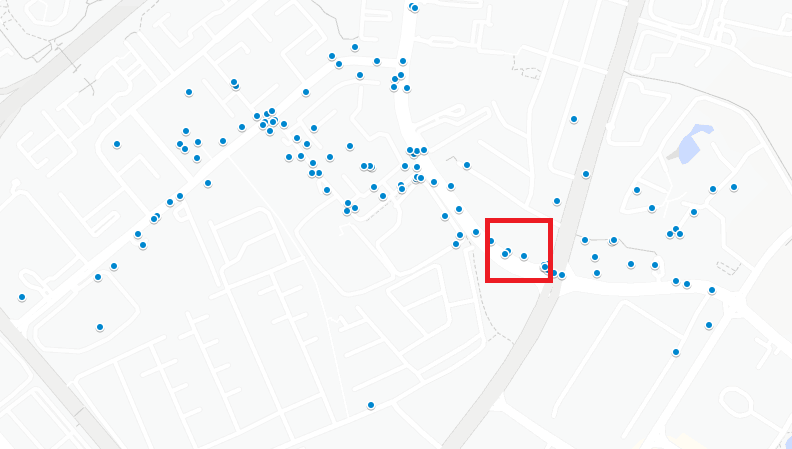}
		\caption{140 GPS points after WiFi based clustering ($\epsilon=0.25$) by all three users.} \label{fig:micro_clustered}
	\end{subfigure}
	\hspace*{\fill} 
	\begin{subfigure}{0.49\textwidth}
		\includegraphics[width=\linewidth]{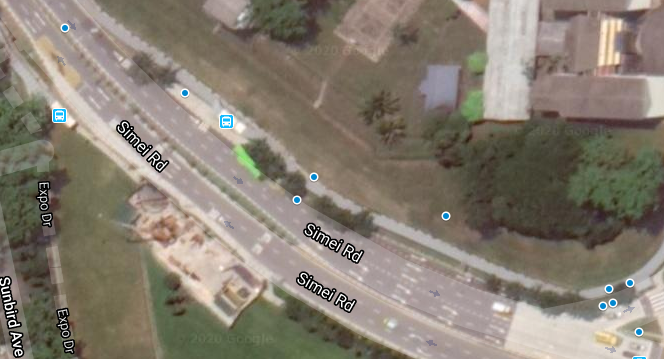}
		\caption{Zoomed in satellite view of GPS points after WiFi based clustering ($\epsilon=0.25$), aligned through the walkway.} \label{fig:satellite_clustered}
	\end{subfigure}	
	\begin{subfigure}{0.49\textwidth}
		\includegraphics[width=\linewidth]{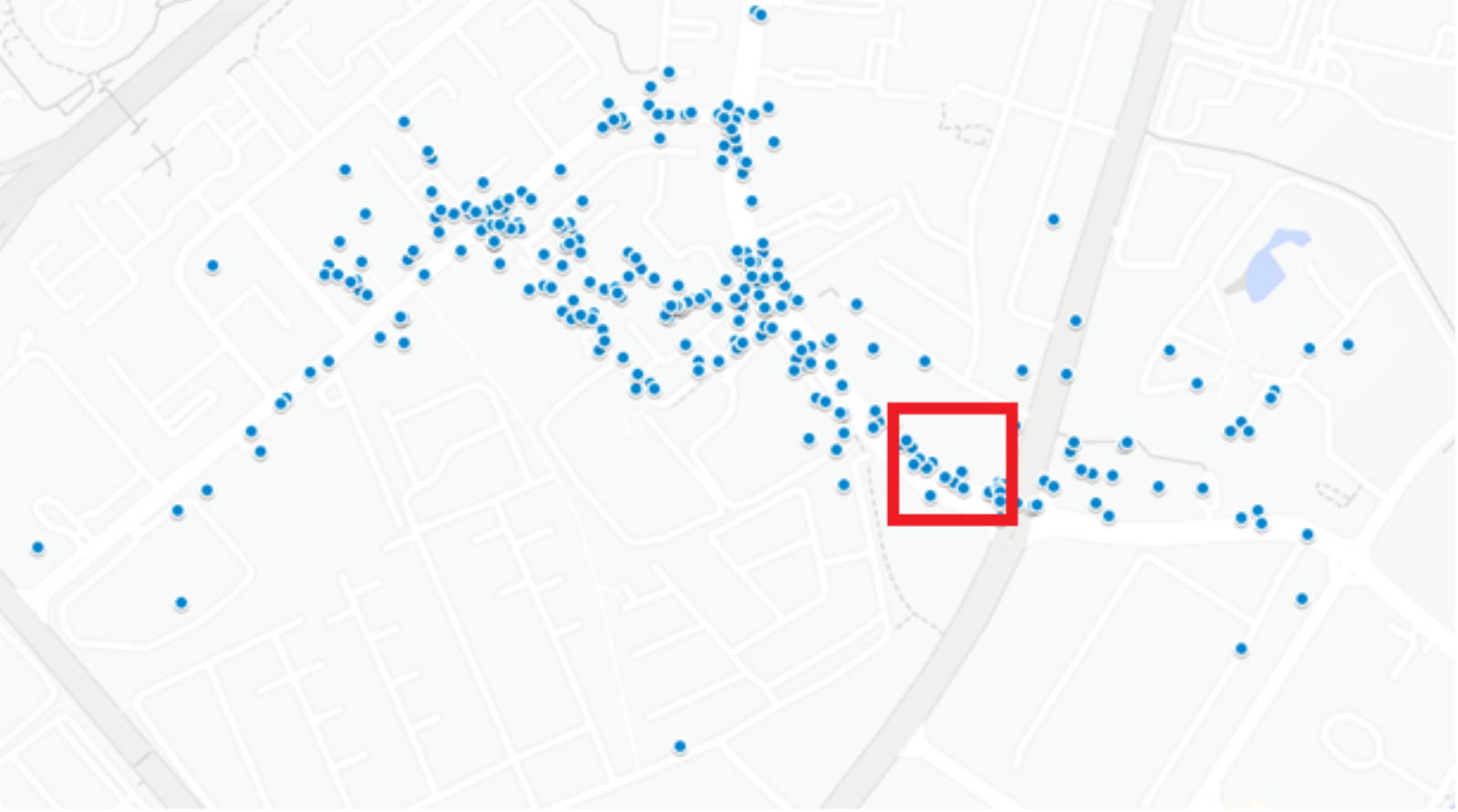}
		\caption{321 GPS points after WiFi based clustering ($\epsilon=0.3$), by all three users.} \label{fig:micro_0.3}
	\end{subfigure}
	\hspace*{\fill} 
	\begin{subfigure}{0.49\textwidth}
		\includegraphics[width=\linewidth]{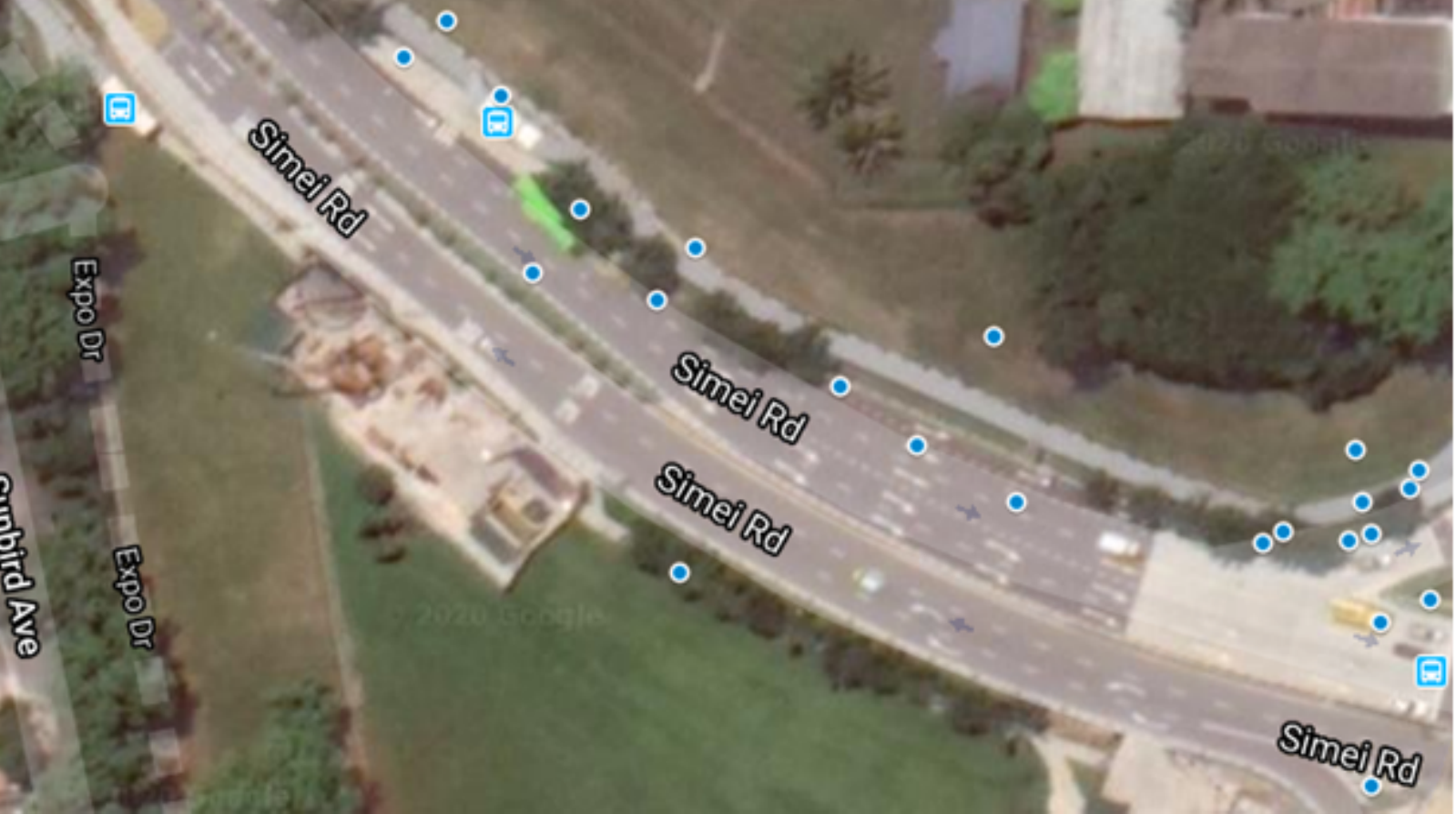}
		\caption{Zoomed in satellite view of GPS points after WiFi based clustering ($\epsilon=0.3$), aligned through the walkway.} \label{fig:satellite_0.3}
	\end{subfigure}		
	\caption{Comparison of before and after WiFi based clustering for different threshold values. 8345 points are reduced into 140 points ($\epsilon=0.25$) and 321 points ($\epsilon=0.3$).} 
	\label{fig:micro_mobility}
\end{figure*}

We study the mobility pattern of $3$ users (denoted as A, B, and L in the Table \ref{tab_smartphones}) from our study, who live in the same neighborhood (i.e. Simei area in Singapore) and work at the same place (i.e. Singapore University of Technology and Design). Most of the times, these $3$ users commute by walking.
The results of the WiFi based clustering is shown in Figure \ref{fig:micro_mobility}.
The figure~\ref{fig:micro_raw} shows the raw GPS for $3$ different users, which consists of GPS data points within the travel duration from individual home to work. 
Note that, each color denotes a separate user L-Purple, A-Yellow, and B-Black, and not all the users have same data amount despite the same timeline, which is from 01 December 2019 to 31 December 2019.

Based on the raw data, we perform data processing as shown in the previous subsection, and cluster the locations, based on WiFi similarities to preserve significant GPS points. A total of $8345$ raw GPS points are simplified into $140$ ($\epsilon=0.25$) and $321$ ($\epsilon=0.3$) clusters as shown in Figure \ref{fig:micro_clustered} and Figure \ref{fig:micro_0.3} respectively, based on WiFi fingerprint clustering method. In other words, each point in Figures \ref{fig:micro_clustered} and \ref{fig:micro_0.3} represents one WiFi based cluster, which is mapped into the nearest GPS point by timestamp. 
By comparing the Figures, we can observe that WiFi based clustering helps to represent the data in a clearer way, instead of the messy data representation obtained by only using GPS data.

Figure \ref{fig:satellite_raw}, Figure \ref{fig:satellite_clustered}, and Figure \ref{fig:satellite_0.3} show zoomed in satellite view of raw gps points shown in the red sqaure area in Figure \ref{fig:micro_raw}, Figure \ref{fig:micro_clustered}, and \ref{fig:micro_0.3} respectively. The red square area has a sheltered walkway at the side of the road. By comparing the figures \ref{fig:satellite_raw}, \ref{fig:satellite_clustered}, and \ref{fig:satellite_0.3}, we can see that when $\epsilon=0.3$, the clustered points are aligned through the walkway. Therefore, WiFi based GPS clustering helps to identify micro mobility patterns of users, which is not possible by only visualizing raw GPS data.

\section{Discussion and Conclusion}
\label{sec_6}

We introduce a mobile crowdsensing system in this paper, to understand three major insights for urban mobility analysis through WiFi fingerprint clustering. 
Data collected from a smartphone application (GPS location, surrounding WiFi access points) are used to identify the indoor POI within a building, obtain neighborhood activity, understand and micro mobility patterns of the users. 

We have demonstrated that, through the fusion of GPS data along with WiFi AP information, it is possible to identify the indoor POI among different users, which are not possible to identify only using GPS location data.
We introduce neighbourhood activity analysis to identify the POI, where users visit for a short break, while staying at home (e.g. visit a common area in the same building, but a different floor level). Since urban apartment complexes are high rise buildings, GPS alone fails to identify such activities, yet the combination of GPS and WiFi can provide meaningful insights.
Also, by such fusion we can identify neighborhood activity. When a user walks under a sheltered walk way, GPS lacks positioning accuracy and fluctuates a lot from the actual physical location. Therefore, it is impossible to capture such mobility patterns by only using GPS stay point extraction. We demonstrated that it is possible to interpret the user mobility paths by WiFi clustering based GPS points, for the purpose of identifying the common trajectories.
For future work, our aim is to deploy the proposed system into a bigger user group, build a POI recommendation platform, and conduct user profiling based on their mobility patterns.

\ifCLASSOPTIONcompsoc
  \section*{Acknowledgments}
\else
  \section*{Acknowledgment}
\fi

This research, led together with the Housing and Development Board, is supported by the Singapore Ministry of National Development and the National Research Foundation, Prime Ministers Office under the Land and Livability National Innovation Challenge (L2 NIC) Research Programme (L2 NIC Award No. L2NICTDF1-2017-4). Any opinions, findings, and conclusions or recommendations expressed in this material are those of the author(s) and do not reflect the views of the Housing and Development Board, Singapore Ministry of National Development and National Research Foundation, Prime Ministers Office, Singapore.

\ifCLASSOPTIONcaptionsoff
  \newpage
\fi



%


\bibliographystyle{IEEEtran}
\bibliography{refer}

\begin{thebibliography}{10}
\providecommand{\url}[1]{#1}
\csname url@samestyle\endcsname
\providecommand{\newblock}{\relax}
\providecommand{\bibinfo}[2]{#2}
\providecommand{\BIBentrySTDinterwordspacing}{\spaceskip=0pt\relax}
\providecommand{\BIBentryALTinterwordstretchfactor}{4}
\providecommand{\BIBentryALTinterwordspacing}{\spaceskip=\fontdimen2\font plus
\BIBentryALTinterwordstretchfactor\fontdimen3\font minus
  \fontdimen4\font\relax}
\providecommand{\BIBforeignlanguage}[2]{{%
\expandafter\ifx\csname l@#1\endcsname\relax
\typeout{** WARNING: IEEEtran.bst: No hyphenation pattern has been}%
\typeout{** loaded for the language `#1'. Using the pattern for}%
\typeout{** the default language instead.}%
\else
\language=\csname l@#1\endcsname
\fi
#2}}
\providecommand{\BIBdecl}{\relax}
\BIBdecl

\bibitem{farkas2015crowdsending}
K.~Farkas, G.~Feher, A.~Benczur, and C.~Sidlo, ``Crowdsending based public
  transport information service in smart cities,'' \emph{IEEE Communications
  Magazine}, vol.~53, no.~8, pp. 158--165, 2015.

\bibitem{leonardi2014secondnose}
C.~Leonardi, A.~Cappellotto, M.~Caraviello, B.~Lepri, and F.~Antonelli,
  ``Secondnose: an air quality mobile crowdsensing system,'' in
  \emph{Proceedings of the 8th Nordic Conference on Human-Computer Interaction:
  Fun, Fast, Foundational}.\hskip 1em plus 0.5em minus 0.4em\relax ACM, 2014,
  pp. 1051--1054.

\bibitem{hu2014multidimensional}
X.~Hu, X.~Li, E.~Ngai, V.~Leung, and P.~Kruchten, ``Multidimensional
  context-aware social network architecture for mobile crowdsensing,''
  \emph{IEEE Communications Magazine}, vol.~52, no.~6, pp. 78--87, 2014.

\bibitem{hoteit2014estimating}
S.~Hoteit, S.~Secci, S.~Sobolevsky, C.~Ratti, and G.~Pujolle, ``Estimating
  human trajectories and hotspots through mobile phone data,'' \emph{Computer
  Networks}, vol.~64, pp. 296--307, 2014.

\bibitem{kang2013exploring}
C.~Kang, S.~Sobolevsky, Y.~Liu, and C.~Ratti, ``Exploring human movements in
  singapore: a comparative analysis based on mobile phone and taxicab usages,''
  in \emph{Proceedings of the 2nd ACM SIGKDD international workshop on urban
  computing}.\hskip 1em plus 0.5em minus 0.4em\relax ACM, 2013, p.~1.

\bibitem{li2008mining}
Q.~Li, Y.~Zheng, X.~Xie, Y.~Chen, W.~Liu, and W.-Y. Ma, ``Mining user
  similarity based on location history,'' in \emph{Proceedings of the 16th ACM
  SIGSPATIAL international conference on Advances in geographic information
  systems}.\hskip 1em plus 0.5em minus 0.4em\relax ACM, 2008, p.~34.

\bibitem{lou2009map}
Y.~Lou, C.~Zhang, Y.~Zheng, X.~Xie, W.~Wang, and Y.~Huang, ``Map-matching for
  low-sampling-rate gps trajectories,'' in \emph{Proceedings of the 17th ACM
  SIGSPATIAL international conference on advances in geographic information
  systems}.\hskip 1em plus 0.5em minus 0.4em\relax ACM, 2009, pp. 352--361.

\bibitem{gamanayake2020cluster}
C.~M. Gamanayake, L.~A. Jayasinghe, B.~Ng, and C.~Yuen, ``Cluster pruning: An
  efficient filter pruning method for edge ai vision applications,'' \emph{IEEE
  Journal of Selected Topics in Signal Processing}, 2020.

\bibitem{helgason2013opportunistic}
{\'O}.~Helgason, S.~T. Kouyoumdjieva, and G.~Karlsson, ``Opportunistic
  communication and human mobility,'' \emph{IEEE Transactions on Mobile
  Computing}, vol.~13, no.~7, pp. 1597--1610, 2013.

\bibitem{suzuki2007learning}
N.~Suzuki, K.~Hirasawa, K.~Tanaka, Y.~Kobayashi, Y.~Sato, and Y.~Fujino,
  ``Learning motion patterns and anomaly detection by human trajectory
  analysis,'' in \emph{Systems, Man and Cybernetics, 2007. ISIC. IEEE
  International Conference on}.\hskip 1em plus 0.5em minus 0.4em\relax IEEE,
  2007, pp. 498--503.

\bibitem{zhou2012iodetector}
P.~Zhou, Y.~Zheng, Z.~Li, M.~Li, and G.~Shen, ``Iodetector: A generic service
  for indoor outdoor detection,'' in \emph{Proceedings of the 10th acm
  conference on embedded network sensor systems}.\hskip 1em plus 0.5em minus
  0.4em\relax ACM, 2012, pp. 113--126.

\bibitem{7917558}
B.~P.~L. Lau, M.~S. Hasala, V.~S. Kadaba, B.~Thirunavukarasu, C.~Yuen, B.~Yuen,
  and R.~Nayak, ``Extracting point of interest and classifying environment for
  low sampling crowd sensing smartphone sensor data,'' in \emph{2017 IEEE
  International Conference on Pervasive Computing and Communications Workshops
  (PerCom Workshops)}, March 2017, pp. 201--206.

\bibitem{Shin2012Unsupervised}
H.~{Shin}, Y.~{Chon}, and H.~{Cha}, ``Unsupervised construction of an indoor
  floor plan using a smartphone,'' \emph{IEEE Transactions on Systems, Man, and
  Cybernetics, Part C (Applications and Reviews)}, vol.~42, no.~6, pp.
  889--898, Nov 2012.

\bibitem{marakkalage2018understanding}
S.~H. Marakkalage, S.~Sarica, B.~P.~L. Lau, S.~K. Viswanath,
  T.~Balasubramaniam, C.~Yuen, B.~Yuen, J.~Luo, and R.~Nayak, ``Understanding
  the lifestyle of older population: Mobile crowdsensing approach,'' \emph{IEEE
  Transactions on Computational Social Systems}, 2018.

\bibitem{marakkalage2019identifying}
S.~H. Marakkalage, R.~Liu, S.~K. Viswanath, and C.~Yuen, ``Identifying indoor
  points of interest via mobile crowdsensing: An experimental study,'' in
  \emph{2019 IEEE VTS Asia Pacific Wireless Communications Symposium
  (APWCS)}.\hskip 1em plus 0.5em minus 0.4em\relax IEEE, 2019, pp. 1--5.

\bibitem{alzantot2012crowdinside}
M.~Alzantot and M.~Youssef, ``Crowdinside: automatic construction of indoor
  floorplans,'' in \emph{Proceedings of the 20th International Conference on
  Advances in Geographic Information Systems}.\hskip 1em plus 0.5em minus
  0.4em\relax ACM, 2012, pp. 99--108.

\bibitem{zhu2014spatio}
J.~Y. Zhu, A.~X. Zheng, J.~Xu, and V.~O. Li, ``Spatio-temporal (st) similarity
  model for constructing wifi-based rssi fingerprinting map for indoor
  localization,'' in \emph{Indoor Positioning and Indoor Navigation (IPIN),
  2014 International Conference on}.\hskip 1em plus 0.5em minus 0.4em\relax
  IEEE, 2014, pp. 678--684.

\bibitem{liu_crowdsensing_2019}
R.~Liu, S.~H. Marakkalage, M.~Padmal, T.~Shaganan, C.~Yuen, Y.~L. Guan, and
  U.-X. Tan, ``Crowd-sensing simultaneous localization and radio fingerprint
  mapping based on probabilistic similarity models,'' in \emph{Proceedings of
  the ION 2019 Pacific PNT Meeting}, Honolulu, Hawaii, April 2019, pp. 73--83.

\bibitem{liu_ieee_sensors2017}
R.~{Liu}, C.~{Yuen}, T.~{Do}, and U.~{Tan}, ``Fusing similarity-based sequence
  and dead reckoning for indoor positioning without training,'' \emph{IEEE
  Sensors Journal}, vol.~17, no.~13, pp. 4197--4207, July 2017.

\bibitem{liu2019collaborative}
R.~Liu, S.~H. Marakkalage, M.~Padmal, T.~Shaganan, C.~Yuen, Y.~L. Guan, and
  U.-X. Tan, ``Collaborative slam based on wifi fingerprint similarity and
  motion information,'' \emph{IEEE Internet of Things Journal}, 2019.

\bibitem{tian2019rf}
X.~Tian, X.~Wu, H.~Li, and X.~Wang, ``Rf fingerprints prediction for cellular
  network positioning: A subspace identification approach,'' \emph{IEEE
  Transactions on Mobile Computing}, vol.~19, no.~2, pp. 450--465, 2019.

\bibitem{ganti2011mobile}
R.~K. Ganti, F.~Ye, and H.~Lei, ``Mobile crowdsensing: current state and future
  challenges,'' \emph{IEEE Communications Magazine}, vol.~49, no.~11, 2011.

\bibitem{lau2019survey}
B.~P.~L. Lau, S.~H. Marakkalage, Y.~Zhou, N.~U. Hassan, C.~Yuen, M.~Zhang, and
  U.-X. Tan, ``A survey of data fusion in smart city applications,''
  \emph{Information Fusion}, vol.~52, pp. 357--374, 2019.

\bibitem{wu2014smartphones}
C.~Wu, Z.~Yang, and Y.~Liu, ``Smartphones based crowdsourcing for indoor
  localization,'' \emph{IEEE Transactions on Mobile Computing}, vol.~14, no.~2,
  pp. 444--457, 2014.

\bibitem{marakkalage2019real}
S.~H. Marakkalage, B.~P.~L. Lau, S.~K. Viswanath, C.~Yuen, and B.~Yuen,
  ``Real-time data analysis using a smartphone mobile application,'' in
  \emph{Ageing and the Built Environment in Singapore}.\hskip 1em plus 0.5em
  minus 0.4em\relax Springer, 2019, pp. 221--240.

\bibitem{wifiscan}
Google, ``{Wi-Fi Scanning},'' \url{https://goo.gl/RqxNk2}, 2018, [Online;
  accessed 01-November-2018].

\bibitem{Zheng2009Mining}
Y.~Zheng, L.~Zhang, X.~Xie, and W.-Y. Ma, ``Mining interesting locations and
  travel sequences from gps trajectories,'' in \emph{Proceedings of the 18th
  International Conference on World Wide Web}, ser. WWW ’09.\hskip 1em plus
  0.5em minus 0.4em\relax New York, NY, USA: Association for Computing
  Machinery, 2009, p. 791–800.

\bibitem{lau2017extracting}
B.~P.~L. Lau, M.~S. Hasala, V.~S. Kadaba, B.~Thirunavukarasu, C.~Yuen, B.~Yuen,
  and R.~Nayak, ``Extracting point of interest and classifying environment for
  low sampling crowd sensing smartphone sensor data,'' in \emph{2017 IEEE
  International Conference on Pervasive Computing and Communications Workshops
  (PerCom Workshops)}.\hskip 1em plus 0.5em minus 0.4em\relax IEEE, 2017, pp.
  201--206.

\bibitem{ester1996density}
M.~Ester, H.-P. Kriegel, J.~Sander, X.~Xu \emph{et~al.}, ``A density-based
  algorithm for discovering clusters in large spatial databases with noise.''
  in \emph{Kdd}, vol.~96, no.~34, 1996, pp. 226--231.

\bibitem{blondel2008fast}
V.~D. Blondel, J.-L. Guillaume, R.~Lambiotte, and E.~Lefebvre, ``Fast unfolding
  of communities in large networks,'' \emph{Journal of statistical mechanics:
  theory and experiment}, vol. 2008, no.~10, p. P10008, 2008.

\end{thebibliography}
%








\end{document}